\documentclass[10pt,journal,compsoc]{IEEEtran}
\ifCLASSOPTIONcompsoc
  \usepackage[nocompress]{cite}
\else
  \usepackage{cite}
\fi

\usepackage[switch]{lineno}
\usepackage[normalem]{ulem}
\usepackage{color}
\usepackage[dvipsnames]{xcolor}
\usepackage{multirow}
\usepackage{graphicx}
\usepackage{float}
\usepackage{amssymb}
\usepackage[font=small]{caption}

\ifCLASSINFOpdf
\else
\fi

\begin{document}
%\linenumbers

\title{Consistent 3D Hand Reconstruction in Video\\ via Self-Supervised Learning}

\author{Zhigang Tu$^\#$,~\IEEEmembership{Member,~IEEE,}
        Zhisheng Huang$^\#$,
        Yujin Chen$^*$, %~\IEEEmembership{Member,~IEEE,} Yujin is not a (student) member
        Di Kang,
        Linchao~Bao,~\IEEEmembership{Member,~IEEE,}
        Bisheng~Yang,~\IEEEmembership{Senior Member,~IEEE,}
        and Junsong Yuan,~\IEEEmembership{Fellow,~IEEE}
\IEEEcompsocitemizethanks{\IEEEcompsocthanksitem Zhigang Tu, Zhisheng Huang, and Bisheng Yang are with the State Key Laboratory of Information Engineering in Surveying, Mapping and Remote Sensing, Wuhan University, Wuhan 430079, China. $^\#$Zhigang~Tu and Zhisheng Huang contributed equally and they are the co-first authors.
\IEEEcompsocthanksitem Yujin Chen is with Technical University of Munich. Work done at Wuhan University. $^*$Correspondence to Yujin Chen (email: yujin.chen@tum.de)
\IEEEcompsocthanksitem Di Kang and Linchao Bao are with Tencent AI Lab.
\IEEEcompsocthanksitem Junsong Yuan is with the Computer Science and Engineering Department, University at Buffalo, Buffalo, NY 14228, USA.}
}

\markboth{IEEE Transactions on Pattern Analysis and Machine Intelligence}%
{Shell \MakeLowercase{\textit{et al.}}: Bare Advanced Demo of IEEEtran.cls for IEEE Computer Society Journals}

\IEEEtitleabstractindextext{%
\begin{abstract}
We present a method for reconstructing accurate and consistent 3D hands from a monocular video. 
We observe that the detected 2D hand keypoints and the image texture provide important cues about the geometry and texture of the 3D hand, which can reduce or even eliminate the requirement on 3D hand annotation. Accordingly, in this work, we propose ${\rm {S}^{2}HAND}$, a self-supervised 3D hand reconstruction model, that can jointly estimate pose, shape, texture, and the camera viewpoint from a single RGB input through the supervision of easily accessible 2D detected keypoints.
We leverage the continuous hand motion information contained in the unlabeled video data and explore ${\rm {S}^{2}HAND(V)}$, which uses a set of weights shared ${\rm {S}^{2}HAND}$ to process each frame and exploits additional motion, texture, and shape consistency constrains to obtain more accurate hand poses, and more consistent shapes and textures.
Experiments on benchmark datasets demonstrate that our self-supervised method produces comparable hand reconstruction performance compared with the recent full-supervised methods in single-frame as input setup, and notably improves the reconstruction accuracy and consistency when using the video training data.

\end{abstract}

\begin{IEEEkeywords}
hand pose estimation, 3D hand reconstruction, video analysis, self-supervision
\end{IEEEkeywords}}

\maketitle

\IEEEdisplaynontitleabstractindextext

% \fi
\IEEEpeerreviewmaketitle

\ifCLASSOPTIONcompsoc
\IEEEraisesectionheading{\section{Introduction}\label{sec:introduction}}
\else

\section{Introduction}
\label{sec:introduction}
\fi

\IEEEPARstart{H}{ands} play a central role in the interaction between humans and the environment, from physical contact and grasping to daily communications via hand gesture.
Learning 3D hand reconstruction is the preliminary step for many computer vision applications such as augmented reality\cite{lee2011two}, sign language translation\cite{camgoz2018neural,camgoz2020sign}, and human-computer interaction\cite{holl2018efficient,parelli2020exploiting,tu2019action}.
However, due to
diverse hand configurations and interaction with the environment,
3D hand reconstruction remains a challenging problem, especially when the task relies on monocular data as input.

\begin{figure}[ht]
\vspace{-1.3cm}
\begin{center}
\includegraphics[width=1\linewidth]{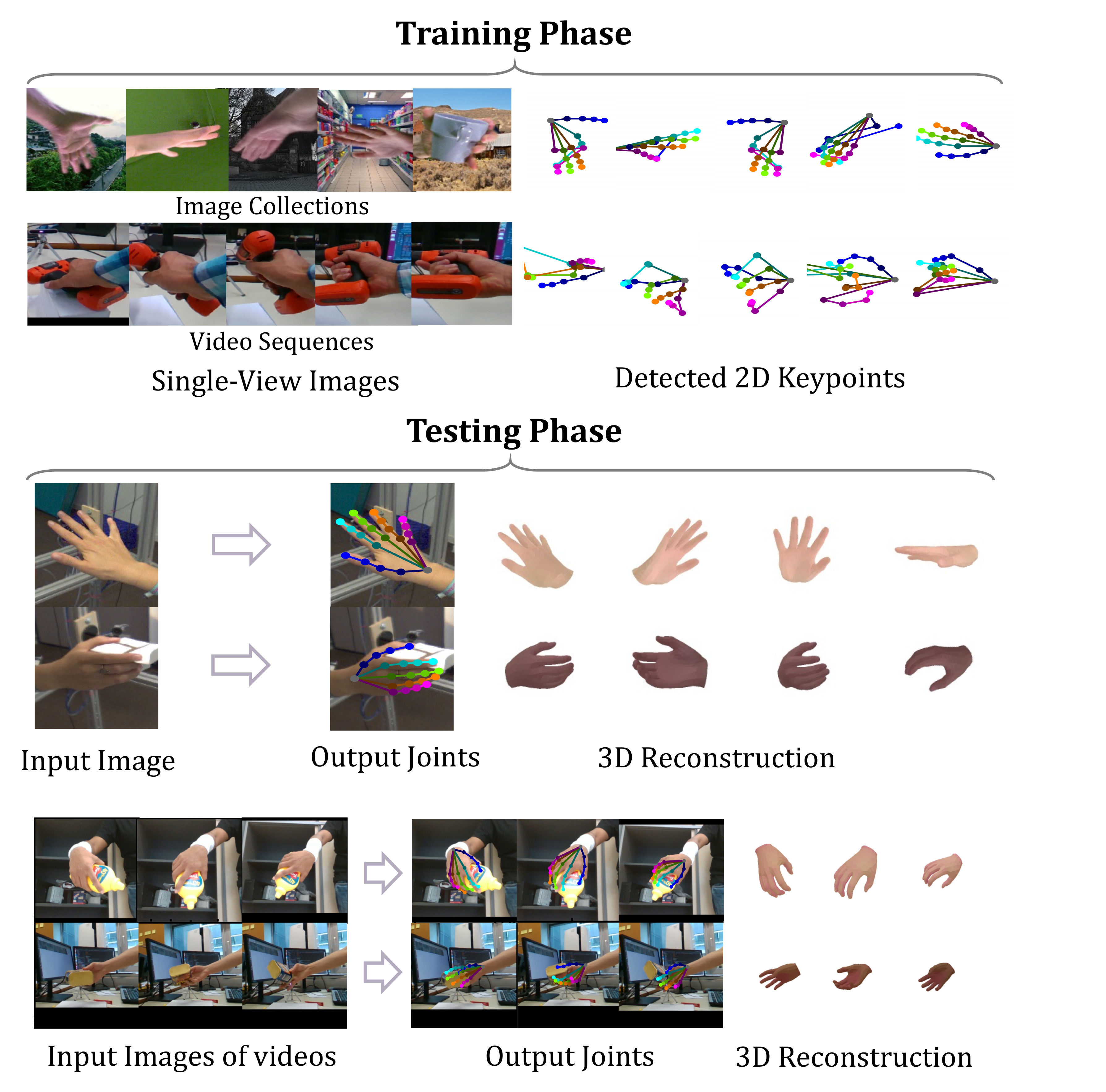}
\caption{Given a collection of unlabeled hand images or videos, we learn a 3D hand reconstruction network in a self-supervised manner.
Top: the training uses unlabeled hand images from image collections or video sequences and their corresponding noisy detected 2D keypoints.
Bottom: our model outputs accurate hand joints and shapes, as well as vivid hand textures.
}
\label{fig:1}
\end{center}
\vspace{-0.8cm}
\end{figure}

Compared with multi-view images \cite{ge2018robust,yu2020humbi,zhao2020hand,poier2018learning} and depth maps \cite{armagan2020measuring,chen2019so,ge2016robust,huang2020hand,yuan2018depth}, the monocular hand images are more common in practice.
In recent years, we have witnessed many efforts in recovering 3D shapes of human hands from single-view RGB images. For example,
\cite{athitsos2003estimating, cai2018weakly,iqbal2018hand,spurr2020weakly,zimmermann2017learning} were proposed to predict 3D hand pose from an RGB image.
However, they only represent the 3D hand through sparse joints, and ignore the 3D shape information, which are required for some applications such as grasping objects with virtual hands \cite{holl2018efficient}.
To better capture the surface information of the hand, previous studies predict the triangle mesh either via regressing per-vertex coordinates \cite{Ge_2019_CVPR,Kulon_2020_CVPR} or by deforming a parametric hand model \cite{hasson2020leveraging,hasson2019learning}. Outputting such high-dimensional representations from 2D input is challenging for neural networks to learn. As a result, the training process relies heavily on 3D hand annotations such as dense hand scans, model-fitted parametric hand mesh, or human-annotated 3D joints. Besides, the hand texture is important in some applications, such as vivid hands reconstruction in immersive virtual reality. But only recent work try to explore parametric texture estimation in a learning-based hand recovery system \cite{qian2020parametric}, while most previous work of 3D hand reconstruction do not consider texture modeling \cite{qian2020parametric}.

One of our key observations is that the 2D cues in the hand image are quite informative to reconstruct the 3D hand model in the real world. The 2D hand keypoints contain rich structural information, and the 2D image contains abundant texture and shape information. Both are important for reducing the use of expensive 3D annotations but have not been fully investigated. Leveraging these cues, we could directly use 2D annotations and the input image to learn the geometry and texture representations without relying on 3D annotations \cite{spurr2020weakly}. However, it is still labor-consuming to annotate 2D hand keypoints per image. To completely save the manual annotation, we propose to extract 2D hand keypoints as well as geometric representations from the unlabeled hand image to help the shape reconstruction and use the texture information contained in the input image to help the texture modeling.

Additionally, video sequences contain rich hand motion and more comprehensive appearance information. Usually, a frame-wise fully-supervised hand reconstruction model does not take these information into serious consideration since 3D annotations already provide a strong supervision. As a result, it is more difficult for a frame-wise model to produce consistent results from video frames compared to sequence-wise models, since no temporal information is utilized. 
Thereby, we propose to penalize the inconsistency of the output hand reconstructions from consecutive observations of the same hand. In this way, motion prior in video is distilled in the frame-wise model to help reconstruct more accurate hand for every single frame. Notably, the constraints on the sequence output are also employed in a self-supervised manner.

Driven by the above observations, this work aims to train an accurate 3D hand reconstruction network using only the supervision signals obtained from the input images or video sequences while eliminating manual annotations of the training images.
To this end, we use an off-the-shelf 2D keypoint detector \cite{cao2019openpose} to generate some noisy 2D keypoints, and supervise the hand reconstruction by these noisy detected 2D keypoints and the input image.
Although our reconstruction network relies on the pre-defined keypoint detector, we call it a self-supervised network, following the naming convention in the face reconstruction literature\cite{chen2020self,tewari2017mofa } as only the self-annotation is provided to the training data.
Further, we leverage the self-supervision signal embedded in the video sequence to help the network produce more accurate and temporally more coherent hand reconstructions.
To learn in a self-supervised manner, there are several issues to be addressed.
First, how to efficiently use joint-wise 2D keypoints to supervise the ill-posed monocular 3D hand reconstruction?
Second, how to handle noise in the 2D detection output since our setting is without utilizing any ground truth annotation?
Third, is it possible to make use of the continuous information contained in video sequences to encourage smoothness and consistency of reconstructed hands in a frame-wise model?

To address the first issue, a model-based autoencoder is learned to estimate 3D joints and shape, where the output 3D joints are projected into 2D image space and forced to align with the detected keypoints during training. However, if we only align keypoints in image space, invalid hand pose often occurs. This may be caused by an invalid 3D hand configuration which is still compatible with the projected 2D keypoints. Furthermore, 2D keypoints cannot reduce the scale ambiguity of the predicted 3D hand. Thus, we propose to learn a series of priors embedded in the model-based hand representations to help the neural network output hand with a reasonable pose and size.

To address the second issue, a trainable 2D keypoint estimator and a novel 2D-3D consistency loss are proposed. The 2D keypoint estimator outputs joint-wise 2D keypoints and the 2D-3D consistency loss links the 2D keypoint estimator and the 3D reconstruction network to make the two mutually beneficial to each other during the training. In addition, we find that the detection accuracy of different samples varies greatly, thus we propose to distinguish each detection item to weigh its supervision strength accordingly.

To address the third issue, we decompose the hand motion into the joint rotations and ensure smooth rotations of hand joints between frames by conforming to a quaternion-based representation. Furthermore, a novel quaternion loss function is proposed to allow all possible rotation speeds. Besides motion consistency, hand appearance is another main concern. A texture and shape (T\&S)  consistency loss function is introduced to regularize the coherence of the output hand texture and shape.

In brief, we present a self-supervised 3D hand reconstruction model ${\rm {S}^{2}HAND}$ and its advanced ${\rm {S}^{2}HAND}(V)$. The models enable us to train neural networks that can predict 3D pose, shape, texture and camera viewpoint from images without any ground truth annotation of training images, except that we use the outputs from a 2D keypoint detector (see Fig. 1). 
Notably, ${\rm {S}^{2}HAND(V)}$ is able to extract informative supervision from unannotated videos to help learn a better frame-wise model. In order to achieve this, ${\rm {S}^{2}HAND(V)}$ inputs the sequential data to multiple weight-shared ${\rm {S}^{2}HAND}$ models and employs proposed constraints on the sequential output at the training stage.

The advantage of our proposed methods are summarized as follows:
\begin{itemize}
    \item We present the first self-supervised 3D hand reconstruction models, which accurately output 3D joints, mesh, and texture from a single image, without using any annotated training data.
    \item We exploit an additional trainable 2D keypoint estimator to boost the 3D reconstruction through a mutual improvement manner, in which a novel 2D-3D consistency loss is constructed.
    \item We introduce a hand texture estimation module to learn vivid hand texture via self-supervision.
    \item We benchmark self-supervised 3D hand reconstruction on some currently challenging datasets, where our self-supervised method achieves comparable performance to previous fully-supervised methods.
\end{itemize}

This work is an extension of our conference paper \cite{chen2021model}. The new contributions include:
\begin{itemize}
    \item We extend our ${\rm {S}^{2}HAND}$ model to the ${\rm {S}^{2}HAND(V)}$ model, which further exploits the self-supervision signals embedded in video sequences. The improvement in accuracy and smoothness is 3.5\% and 3.1\%, respectively.
    \item We present a quaternion loss function, which is based on an explored motion-aware joints rotation representation, to help learn smooth hand motion. Experiments demonstrate its significant advantage over the similar methods in both accuracy and smoothness.
    \item We propose a texture and shape consistency regularization term to encourage coherent shape and texture reconstruction.
    \item We illustrate that utilizing extra in-the-wild unlabeled training data can further boost the performance of our model.
\end{itemize}
%\vspace{-0.5cm}

\section{Related Work}
\label{sec:related_work}
\textbf{\textit{Hand Pose and Shape Estimation}}.
Researchers have developed a lot of different methods in hand pose and shape estimation, such as regression-based method \cite{cai20203d,zimmermann2017learning,yang2019disentangling,lin2021end,chen2021i2uv} and model-based method \cite{liu2021semi,zhou2021monocular,chen2021camera,cao2021reconstructing}.
Comparing to hand pose which is represented by 3D coordinates of hand joints alone, hand mesh contains more detailed shape information and recently has become the focus in the research community. Several methods utilize the hand mesh topology to directly output 3D mesh vertices. E.g. \cite{kulon2020weakly,lim2018simple,dkulon2019rec} use the spiral convolution to recover hand mesh and \cite{defferrard2016convolutional,ge20193d,lin2021end} use the graph convolution to output mesh vertices.
%Though such methods introduces as little prior as possible, they need lots of annotated data for training in exchange.
Although these methods introduce as few priors as possible, they require large amounts of annotated data for training.
In this self-supervised work, we make use of the priors contained in the MANO hand model \cite{MANO:SIGGRAPHASIA:2017}, where MANO can map pose and shape parameters to a triangle mesh \cite{boukhayma20193d,chen2021joint,hasson2019learning, zimmermann2019freihand}, to reduce reliance on the labeled training data.%\dknote{add ref: other paper using MANO model or mention it's widely adopted by many papers. Otherwise, it sounds like nobody else has used it for this task.}

Because the parametric model contains abundant structure priors of human hands, recent works integrate hand model as a differentiable layer in neural networks \cite{Baek_2019_CVPR,baek2020weakly,boukhayma20193d,hasson2020leveraging,hasson2019learning,wang_SIGAsia2020,zhou2020monocular,zimmermann2019freihand}.
Among them, \cite{Baek_2019_CVPR,wang_SIGAsia2020, zhou2020monocular} output a set of intermediate estimations, like segmentation mask and 2D keypoints, and then map these representations to the MANO parameters.
Different from them, we aim at demonstrating the feasibility of a self-supervised framework using an intuitive autoencoder. We additionally output 2D keypoint estimation from another branch and utilize it only during training to facilitate 3D reconstruction.
More generally, recent methods \cite{baek2020weakly, boukhayma20193d,hasson2020leveraging,hasson2019learning, zimmermann2019freihand} directly adopt an autoencoder that couples an image feature encoding stage with a model-based decoding stage.
Unlike \cite{hasson2020leveraging,hasson2019learning}, we focus on hand recovery and do not use any annotation about objects.
More importantly, the above methods use 3D annotations as supervision, while the proposed method does not rely on any ground truth annotations.

\textbf{\textit{3D Hand Pose and Shape Estimation with Limited Supervision}}.
2D annotation is cheaper than 3D annotation, but it is difficult to deal with the ambiguity of depth and scale.
\cite{cai2018weakly} uses a depth map to perform additional weak supervision to strengthen 2D supervision.
\cite{spurr2020weakly} proposes the biomechanical constraints to help the network output feasible 3D hand configurations.
\cite{panteleris2018using} detects 2D hand keypoints and directly fits a hand model to the 2D detection.
\cite{Kulon_2020_CVPR} gathers a large-scale dataset through an automated data collection method similar to \cite{panteleris2018using} and then applies the collected mesh as supervision.
In this work, we limit biomechanical feasibility by introducing a set of constraints on the skin model instead of only imposing constraints on the skeleton as~\cite{spurr2020weakly}.
In contrast to \cite{cai2018weakly, Kulon_2020_CVPR}, our method is designed to verify the feasibility of (noisy) 2D supervision and avoids introducing any extra 2.5D or 3D data.

\begin{figure*}[t]
\vspace{-0.3cm}
\centering
\includegraphics[width=\linewidth]{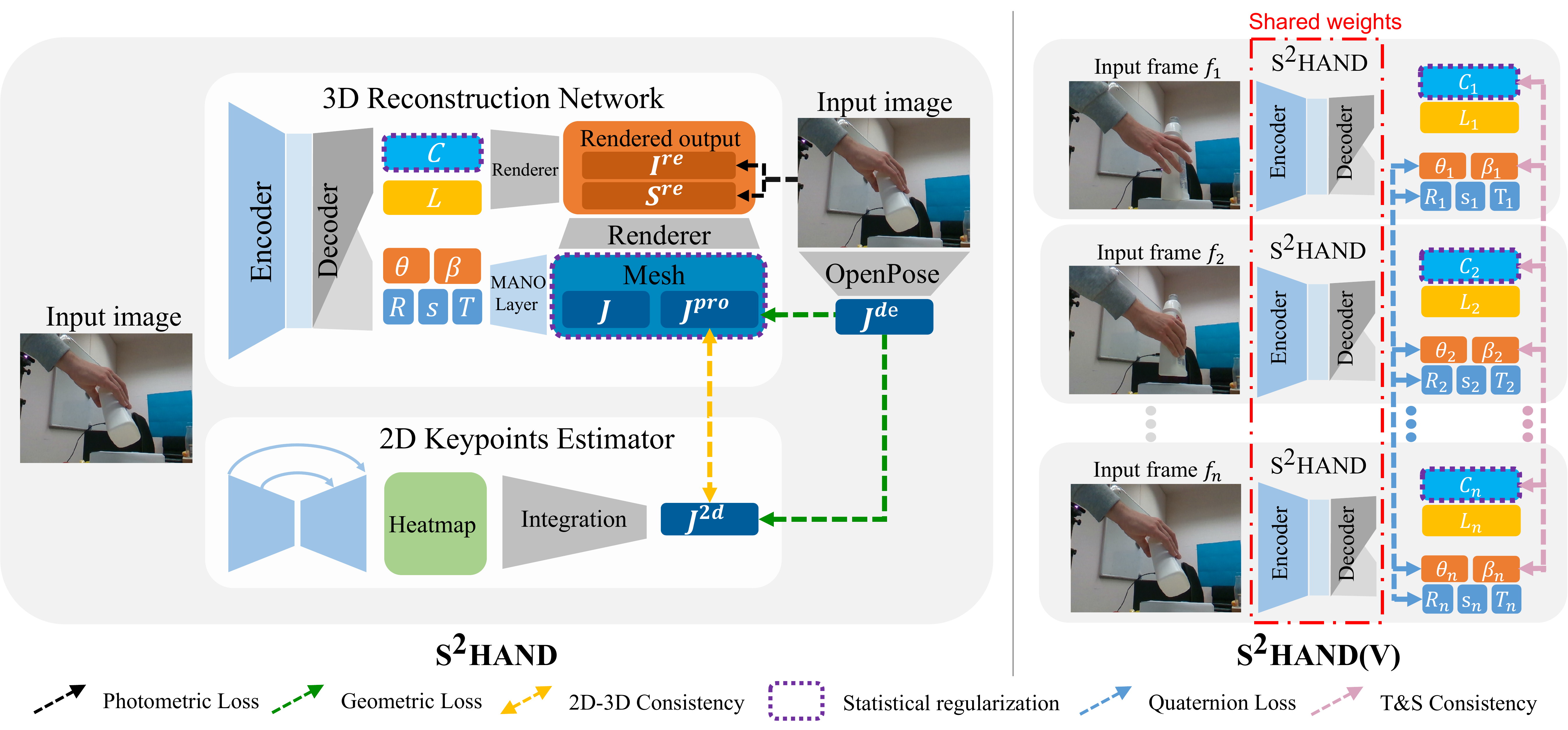}
\caption{
Overview of the proposed models. The ${\rm {S}^{2}HAND(V)}$ on the right learns to reconstruct consistent 3D hands from video sequences without ground truth annotations based on ${\rm {S}^{2}HAND}$.
Given an input image, the ${\rm {S}^{2}HAND}$ model generates a 3D textured hand with its corresponding multiple 2D representations through a 3D reconstruction network and a 2D keypoints estimator.
Effective loss functions and regularization terms are designed for self-supervised network training.
Given a video sequence, the ${\rm {S}^{2}HAND(V)}$ model produces sequential outputs from several weight-shared ${\rm {S}^{2}HAND}$ models with temporal constraints.
A quaternion loss and a T\&S loss are presented to exploit continuous motion information to promote consistent hand reconstruction.
During the inference, only the 3D reconstruction network is utilized and the ${\rm {S}^{2}HAND(V)}$ acts just like a specially trained ${\rm {S}^{2}HAND}$ due to weight sharing.
The symbols used in this figure can be found in Section~\ref{section:sshrfimc} and Section~\ref{section:sshrfvq}.
}
\label{fig:pipeline}
%\label{fig:onecol}
\vspace{-0.5cm}
\end{figure*}

\textbf{\textit{Self-supervised 3D Reconstruction}}.
Recently, there are methods that propose to learn 3D geometry from the monocular image only.
For example, \cite{Wu_2020_CVPR} presents an unsupervised approach to learn 3D deformable objects from raw single-view images, but they assume the object is perfectly symmetric, which is not the case in the hand reconstruction.
\cite{goel2020shape} removes keypoints from the supervision signals, but it uses ground truth 2D silhouette as supervision and only tackles categories with small intra-class shape differences, such as birds, shoes, and cars.\cite{spurr2021self} exploits a self-supervised contrastive learning for hand pose estimation, but only the encoder is pretrained in the self-supervised manner. \cite{guo2020graph} designs a self-supervised module to overcome inconsistency between the 2D and 3D hand pose, but they only consider the sparse joint keypoints.
\cite{wan2019self} explores a depth-based self-supervised 3D hand pose estimation method, but the depth image provides much stronger evidence and supervision than the RGB image.
Recently, \cite{chen2020self, tewari2018self, tewari2017mofa} exploits a self-supervised face reconstruction method with the usage of 3D morphable model of face (3DMM) \cite{blanz1999morphable} and 2D landmarks detection.
Our approach is similar to them, but the hand is non-flat and asymmetrical when compared with the 3D face, and the hand suffers from more severe self-occlusion. These characteristics make this self-supervised hand reconstruction task more challenging.

\textbf{\textit{Texture Modeling in Hand Recovery}}.
\cite{de2011model, de2008model} exploit shading and texture information to handle the self-occlusion problem in the hand tracking system. Recently, \cite{qian2020parametric} uses principal component analysis (PCA) to build a parametric texture model of hand from a set of textured scans. In this work, we try to model texture from self-supervised training without introducing extra data, and further investigate whether the texture modeling helps with the shape modeling.

\textbf{\textit{Motion Learning from Sequence Data for 3D Hand Estimation}}.
To leverage motion information contained in sequence data, several methods have been proposed in hand pose estimation. \cite{hasson2020leveraging} uses the photometric consistency between neighboring frames of sparsely annotated RGB videos. \cite{cai2019exploiting} presents a graph-based method to exploit spatial and temporal relationship for sequence pose estimation.\cite{chen2021temporal} utilizes the temporal information through bidirectional inferences.
\cite{yang2020seqhand,liu2021semi} design a temporal consistency loss for motion smoothness. However, these methods either are specialized for motion generation or only impose a weak regularization for motion smoothness.
 
There exists no approach to capture hand motion dynamics fundamentally, leading to limited benefits can be gained from modeling motion. In this work, we aim to exploit self-supervised information from hand motion dynamics.
Unlike most of the previous approaches \cite{fragkiadaki2015recurrent,martinez2017human,aksan2019structured,hernandez2019human} which adopt recurrent or graph-based network structure to learn hand motion in a sequence-to-sequence manner, we instead use a motion-related loss function to help our frame-wise model converges better and bridges the gap with fully-supervised methods.

From the above analysis and comparison, we believe that self-supervised 3D hand reconstruction is feasible and significant, but to the best of our knowledge, no such idea has been studied in this field.
In this work, we fill this gap and propose the first self-supervised 3D hand reconstruction model, and prove its effectiveness through extensive experiments.
\vspace{-0.2cm}

\section{Methodology}
\subsection{Overview}
Our method enables end-to-end learning of accurate and consistent 3D hand reconstruction from video sequences in a self-supervised manner through ${\rm {S}^{2}HAND(V)}$ (Section~\ref{section:sshrfvq}), which is based on ${\rm {S}^{2}HAND}$ (Section~\ref{section:sshrfimc}). The overview is illustrated in Fig.~\ref{fig:pipeline}.

The ${\rm {S}^{2}HAND}$ model takes an image as input and generates a textured 3D hand represented by pose, shape and texture, along with corresponding lighting, camera viewpoint (Section~\ref{sec:encoding} and \ref{sec:decoding}) and multiple 2D representations in the image space (Section~\ref{sec:2D}).
Some efficient loss functions and regularization terms (Section~\ref{image TO}) are explored to train the network without using ground truth annotations.
The ${\rm {S}^{2}HAND(V)}$ model takes video sequences as input and produces consistent sequential outputs from multiple ${\rm {S}^{2}HAND}$ models where their weights are shared.
%with motion-related constraints (Section~\ref{sq TO}).
A quaternion loss (Section~\ref{Sql}) and a T\&S consistency loss (Section~\ref{CT&S}) are designed to train the network with temporal constraints.
We describe the proposed method in detail as below.

\subsection{Self-supervised Hand Reconstruction from Image Collections}
\label{section:sshrfimc}
The ${\rm {S}^{2}HAND}$ model learns self-supervised 3D hand reconstruction from image collections via training a 3D hand reconstruction network with the help of a trainable 2D keypoints estimator (See Section~\ref{sec:2D}).

\subsubsection{Deep Hand Encoding}\label{sec:encoding}
Given an image $I$ that contains a hand, the 3D hand reconstruction network first extracts the feature maps with the EfficientNet-b0 backbone \cite{tan2019efficientnet}, and then transforms them into a geometry semantic code vector $x$ and a texture semantic code vector $y$.
The geometry semantic code vector $x$ parameterizes the hand pose $\theta \in \mathbb{R}^{30}$, shape $\beta \in \mathbb{R}^{10}$, scale $s \in \mathbb{R}^{1}$, rotation $R \in \mathbb{R}^{3}$ and translation $T \in \mathbb{R}^{3}$ in a unified manner: $x = (\theta, \beta, s, R, T)$. The texture semantic code vector $y$ parameterizes the hand texture $C \in \mathbb{R}^{778\times3}$ and scene lighting $L \in \mathbb{R}^{11}$ in a unified manner: $y = (C, L)$.

\subsubsection{Model-based Hand Decoding}\label{sec:decoding}
Given the geometry semantic code vector $x$ and the texture semantic code vector $y$, our model-based decoder generates a textured 3D hand model in the camera space.
In the following, we will describe the used hand model and decoding network in detail.\\
\textbf{Pose and Shape Representation.}\quad The hand surface is represented by a manifold triangle mesh $M\equiv(V, F)$ with $n=778$ vertices $V=\lbrace v_{i}\in \mathbb{R}^{3} \vert 1\le i \le n\rbrace$ and faces $F$.
The faces $F$ indicates the connection of the vertices in the hand surface, where we assume the face topology keeps fixed.
Given the mesh topology, a set of $k=21$ joints \cite{hasson2019learning} $J=\lbrace j_{i}\in \mathbb{R}^{3} \vert 1\le i \le k\rbrace$ (as shown in Fig.~\ref{fig:prior}A) can be directly formulated from the hand mesh.
Here, the hand mesh and joints are recovered from the pose vector $\theta$ and the shape vector $\beta$ via MANO, where MANO is a low-dimensional parametric model \cite{MANO:SIGGRAPHASIA:2017}.\\
\textbf{3D Hand in Camera Space.}\quad After representing 3D hand via MANO hand model from pose and shape parameters, the mesh and joints are located in the hand-relative coordinate systems.
To represent the output joints and mesh in the camera coordinate system, we use the estimated scale, rotation and translation to conserve the original hand mesh $M_{0}$ and joints $J_{0}$ into the final representations in terms of: $M = s{M}_{0}R + T$ and $J = s{J}_{0}R + T$.
%{$M$, $J$}:
%\begin{equation}M = s{M}_{0}R + T\end{equation}
%\begin{equation}J = s{J}_{0}R + T\end{equation}
\\
\textbf{Texture and Lighting Representation.}\quad We use per-face RGB value of $1538$ faces to represent the texture of hand $C = \lbrace c_{i} \in \mathbb{R}^{3}| 1\le i \le n \rbrace$, where $c_{i}$ yields the RGB values of vertex $i$.
In our model, we use a simple ambient light and a directional light to simulate the lighting conditions \cite{kato2018neural}.
The lighting vector $L$ parameterizes ambient light intensity $l^a \in \mathbb{R}^{1}$, ambient light color $l^a_c \in \mathbb{R}^{3}$, directional light intensity $l^d \in \mathbb{R}^{1}$, directional light color $l^d_c \in \mathbb{R}^{3}$, and directional light direction $n^d \in \mathbb{R}^{3}$ in a unified representation: $L = (l^a, l^a_c, l^d, l^d_c, n^d)$.
%\begin{equation}L = (l^a, l^a_c, l^d, l^d_c, n^d)\end{equation}}

\subsubsection{2D Hand Representations}\label{sec:2D}
A set of estimated 3D joints within the camera space can be projected into the image space by camera projection.
Similarly, the output textured model can be formulated into a realistic 2D hand image through a neural renderer.
In addition to the 2D keypoints projected from the model-based 3D joints, we can also estimate the 2D position of each keypoint in the input image. Here, we represent 2D hand with three modes and explore the complementarity among them.

\noindent \textbf{Joints Projection.}\quad Given a set of 3D joints in camera coordinates $J$ and the intrinsic parameters of the camera, we use perspective camera projection $\Pi$ to project 3D joints into a set of $k=21$ 2D joints $J^{pro}=\lbrace j^{pro}_i \in \mathbb{R}^{2} \vert 1\le i \le k\rbrace$, where $j^{pro}_i$ yields the position of the $i$-th joint in image UV coordinates: $J^{pro} = \Pi (J)$.

\noindent \textbf{Image Formation.}\quad A 3D mesh renderer is used to conserve the triangle hand mesh into a 2D image, here we use the implementation\footnote{\scriptsize
{https://github.com/daniilidis-group/neural\_renderer}} of \cite{kato2018neural}.
Given the 3D mesh $M$, the texture of the mesh $C$ and the lighting $L$, the neural renderer~$\Delta$ can generate a silhouette of hand $S^{re}$ and a color image $I^{re}$: $S^{re},I^{re}  = \Delta (M,C,L)$.\\% and a depth map $D^{re}$:  \Re
%\begin{equation}S^{re},I^{re}  = \Delta (M,C,L)\end{equation}\\
\noindent \textbf{Extra 2D Joint Estimation.}\quad
Projecting model-based 3D joints into 2D can help the projected 2D keypoints retain the structural information, but at the same time gives up the knowledge of per-joint prior. 
To address this issue, we additionally use a 2D keypoint estimator to directly estimate a set of $k=21$ independent 2D joints \mbox{$J^{2d} =\lbrace j^{2d}_i \in \mathbb{R}^{2} \vert 1\le i \le k\rbrace$}, where $j^{2d}_i$ indicates the position of the $i$-th joint in image UV coordinates. In our 2D keypoint estimator, a stacked hourglass network \cite{newell2016stacked} along with an integral pose regression \cite{sun2018integral} is used. Note that the 2D hand pose estimation module is optionally deployed in the training period and is not required during the inference.

\subsubsection{Training Objective}
\label{image TO}
Our overall training loss $E_{S^{2}HAND}$ consists of three parts, i.e. a 3D branch loss $E_{3d}$, a 2D branch loss $E_{2d}$ and a 2D-3D consistency loss $E_{cons}$:
\begin{equation}E_{S^{2}HAND}= w_{3d}E_{3d} + w_{2d}E_{2d} + w_{cons}E_{cons}\end{equation}
Note, $E_{2d}$ and $E_{cons}$ are optional and only used when the 2D estimator is applied. The constant weights $w_{3d}$, $w_{2d}$ and $w_{cons}$ balance the three terms. In the following, we describe these loss terms in detail.

To train the model-based 3D hand decoder, we enforce geometric alignment $E_{geo}$, photometric alignment $E_{photo}$, and statistical regularization $E_{regu}$:
\begin{equation}
    E_{3d}  = w_{geo}E_{geo} + w_{photo}E_{photo} + w_{regu}E_{regu}
\end{equation}
\textbf{Geometric Alignment.}\quad We propose a geometric alignment loss $E_{geo}$ based on the detected 2D keypoints which are obtained through an implementation\footnote{\scriptsize {https://github.com/Hzzone/pytorch-openpose}} of~\cite{cao2019openpose}.
The detected 2D keypoints $L =\lbrace (j^{de}_{i}, {con}_{i}) \vert 1\le i \le k \rbrace$ allocate each keypoint with a 2D position $j^{de}_{i} \in \mathbb{R}^{2}$ and a 1D confidence ${con}_{i} \in [0,1]$.
The geometric alignment loss in the 2D image space consists of a joint location loss $E_{loc}$ and a bone orientation loss $E_{ori}$.
The joint location loss $E_{loc}$ enforces the projected 2D keypoints $J^{pro}$ to be close to its corresponding 2D detections $J^{de}$, and the bone orientation loss $E_{ori}$ enforces the $m=20$ bones of the keypoints in these two sets to be aligned:
\begin{equation}
    E_{loc}=\frac{1}{k}\sum_{i=1}^{k}{{con}_i}\mathcal{L}_{SmoothL1}(j^{de}_{i},j^{pro}_{i})
\label{eq:lmloc}\end{equation}
\begin{equation}
    E_{ori}=\frac{1}{m}\sum_{i=1}^{m}{{conf}^{bone}_i}{\parallel \nu^{de}_{i}-\nu^{pro}_{i}\parallel}^2_2
\label{eq:lmori}
\end{equation}
Here, a SmoothL1 loss \cite{huber1992robust} is used in Eq.~\ref{eq:lmloc} to make the loss term to be more robust to local adjustment since the detected keypoints are not fit well with the MANO keypoints.
In~Eq.~\ref{eq:lmori}, $\nu^{de}_{i}$ and $\nu^{pro}_{i}$ are the normalized $i$-th bone vector of the detected 2D joints and the projected 2D joints, respectively, and ${conf}^{bone}_i$ is the product of the confidence of the two detected 2D joints of the $i$-th bone. The overall geometric alignment loss $E_{geo}$ is the weighted sum of $E_{loc}$ and $E_{ori}$ with a weighting factor $w_{ori}$:
\begin{equation}
\label{eq:geo_s}
    E_{geo} = E_{loc} + w_{ori}E_{ori}
\end{equation}

\noindent \textbf{Photometric Consistency.}\quad For the image formation, the ideal result is the rendered color image $I^{re}$ matches the foreground hand of the input $I$. To this end, we employ a photometric consistency which has two parts: the pixel loss ${E}_{pixel}$ is computed by averaging the least absolute deviation (L1) distance for all visible pixels to measure the pixel-wise difference, and the structural similarity (SSIM) loss ${E}_{SSIM}$ is estimated by evaluating the structural similarity between the two images \cite{wang2004image}:
\begin{equation}
\label{eq:pho}
{E}_{pixel}=\frac{{conf}_{sum}}{\mid S^{re}\mid}\sum_{(u,v)\in S^{re}}{\parallel I_{u,v}-I^{re}_{u,v} \parallel}_1
\end{equation}
\begin{equation}
{E}_{SSIM}=1-SSIM({I}\odot{S^{re}},I^{re})
\end{equation}
Here, the rendered silhouette $S^{re}$ is used to get the foreground part of the input image for loss computation.
In Eq.~\ref{eq:pho}, we use ${conf}_{sum}$, which is the sum of the detection confidence of all keypoints, to distinguish different training samples, such as occluded ones.
This is because we think that low-confidence samples correspond to ambiguous texture confidence, e.g., the detection confidence of an occluded hand is usually low.
The photometric consistency loss ${E}_{photo}$ is the weighted sum of ${E}_{pixel}$ and ${E}_{SSIM}$ with a weighting factor $w_{SSIM}$:
\begin{equation}
{E}_{photo}={E}_{pixel}+w_{SSIM}{E}_{SSIM}
\vspace{-0.4cm}
\end{equation}
\begin{figure}[tb]
\begin{center}
		%\fbox{\rule{0pt}{2in} \rule{0.7\linewidth}{0pt}}
\includegraphics[width=1\linewidth]{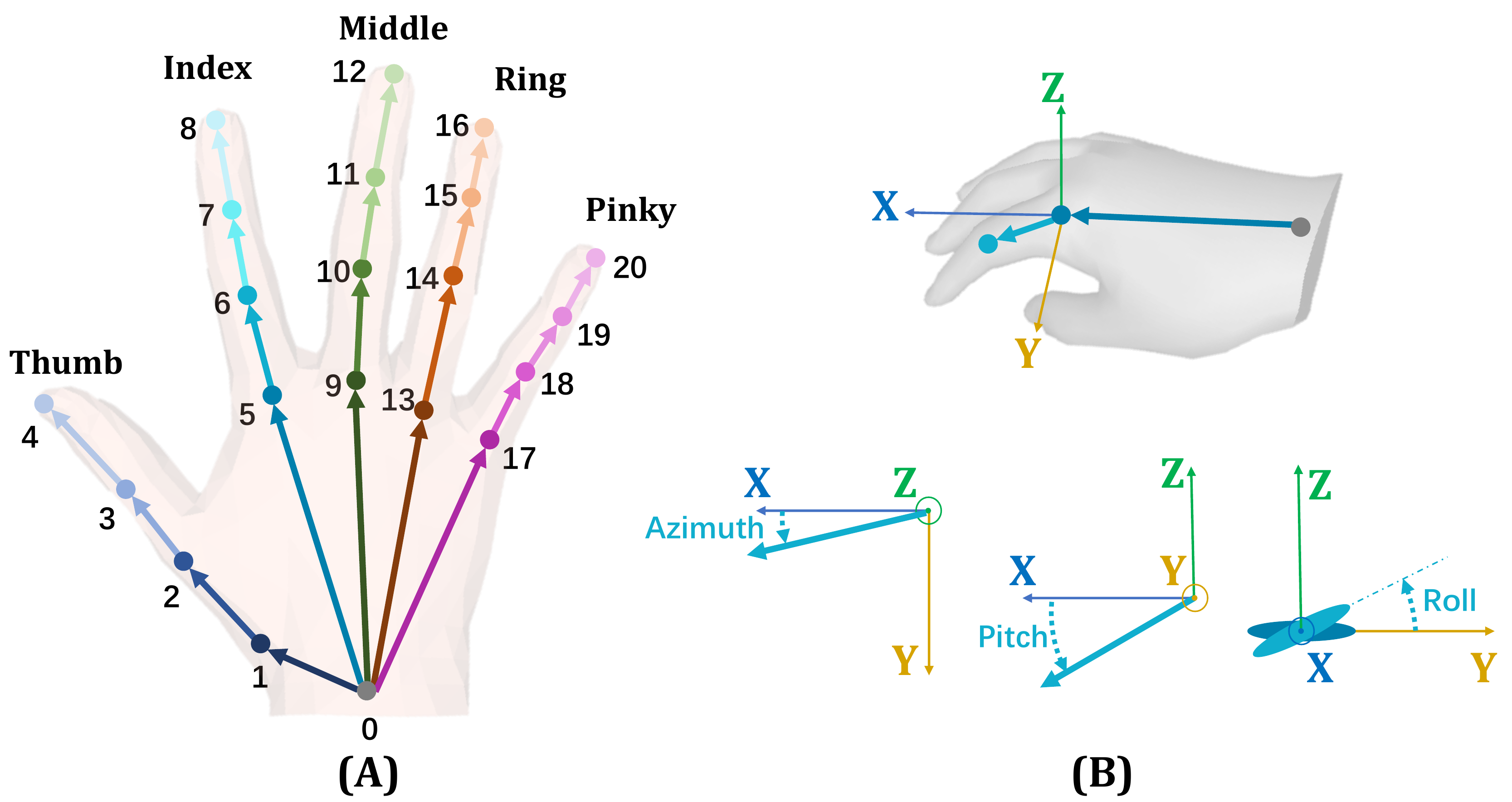}
\end{center}
\caption{(A)The joint skeleton structure. (B) A sample of bone rotation angles. The five bones ($\overrightarrow{0\underline{1}}$,$\overrightarrow{0\underline{5}}$,$\overrightarrow{0\underline{9}}$,$\overrightarrow{0\underline{13}}$,$\overrightarrow{0\underline{17}}$) on the palm are fixed.
Each finger has 3 bones, and the relative orientation of each bone from its root bone is represented by azimuth, pitch, and roll.}
\label{fig:prior}
\vspace{-0.5cm}
\end{figure}

\noindent \textbf{Statistical Regularization.}\quad
During training, to make the results plausible, we introduce some regularization terms, including the shape regularization $E_{\beta}$, the texture regularization $E_{C}$, the scale regularization $E_{s}$, and the 3D joints regularization $E_{J}$. The shape regularization term is defined as $E_{\beta}=\parallel \beta-\bar{\beta} \parallel$ to encourage the estimated hand model shape $\beta$ to be close to the average shape $\bar{\beta}= \vec{0} \in \mathbb{R}^{10}$.
The texture regularization $E_{C}$ is used to penalize outlier RGB values.
The scale regularization term $E_{s}$ is used to ensure the output hand has appropriate size, so as to help determining the depth of the output in this monocular 3D reconstruction task.
To enforce the regularizations on skeleton $E_{J}$, we define feasible range for each rotation angle $a_i$ (as shown in Fig.~\ref{fig:prior}B) and penalize those who exceed the feasible threshold.
The remaining $E_{C}$, $E_{s}$ and $E_{J}$ terms follow \cite{chen2021model}.

The statistical regularization $E_{regu}$ is the weighted sum of $E_{\beta}$, $E_{C}$, ${E}_{s}$ and $E_{J}$ with weighting factors $w_{C}$, $w_{s}$ and $w_{J}$:
\begin{equation}
{E}_{regu}={E}_{\beta}+w_{C}{E}_{C}+w_{s}{E}_{s}+w_{J}{E}_{J}
\end{equation}
\textbf{2D Branch Loss.}\quad For the 2D keypoint estimator, we use a joint location loss as in Eq.~\ref{eq:lmloc} with replacing the projected 2D joint $j_i^{pro}$ by the estimated 2D joint $j_i^{2d}$:
\begin{equation}
    E_{2d}=\frac{1}{k}\sum_{i=1}^{k}{{con}_i}\mathcal{L}_{SmoothL1}(j^{de}_{i},j^{2d}_{i})
\label{eq:lm2d}
\end{equation}
\textbf{2D-3D Consistency Loss.}\quad Since the outputs of the 2D branch and the 3D branch are intended to represent the same hand in different spaces, they should be consistent when they are transferred to the same domain.
Through this consistency, structural information contained in the 3D reconstruction network can be introduced into the 2D keypoint estimator, and meanwhile the estimated 2D keypoints can provide joint-wise geometric cues for 3D hand reconstruction.
To this end, we propose a novel 2D-3D consistency loss to link per projected 2D joint $j_i^{pro}$ with its corresponding estimated 2D joint $j_{i}^{2d}$:
\begin{equation}
    E_{cons}=\frac{1}{k}\sum_{i=1}^{k}\mathcal{L}_{SmoothL1}(j^{pro}_{i},j^{2d}_{i})
\label{eq:cons}
\end{equation}

\subsection{Consistent Self-supervised Hand Reconstruction from Video Sequences}
\label{section:sshrfvq}
The ${\rm {S}^{2}HAND(V)}$ model learns consistent self-supervised 3D hand reconstruction from video sequences via training weight-shared ${\rm {S}^{2}HAND}$ models with temporal constraints, including a quaternion loss and a T\&S consistency loss.

%\subsubsection{Quaternion loss}
\subsubsection{Quaternion-based Motion Regularization}
\label{Sql}
We reformulate hand motion in joint rotation perspective.
We choose the unit quaternion\cite{pavllo2020modeling,zhang2020quaternion,shoemake1985animating,zhou2019continuity} as our joint rotation representation, which can represent spatial orientations and rotations of elements in a convenient and efficient way.
The unit quaternion associated with a spatial rotation is constructed as:
\begin{equation}
    \mathbf{q} = \left ( \cos(\frac{\alpha }{2}), \sin(\frac{\alpha }{2})\vec{\mathbf{u}} \right)
\label{eq:quat_rot}
\end{equation}
where $\alpha$ is the rotation angle and $\vec{\mathbf{u}}$ denotes the rotation axis in $\mathbb{R}^{3}$. Notably, $\mathbf{q}$ can represent both rotation and orientation.

Smooth orientation transition $\mathbf{q}_{t}$ between initial $\mathbf{q}_{0}$ joint orientation and final $\mathbf{q}_{1}$ joint orientation is defined by a unique axis $\vec{\mathbf{v}}$ and corresponding rotation angle $\mathbf{\gamma}$ around the axis. The transition process can be expressed as follows:
\begin{equation}
     \mathbf{q}_{t} = (\mathbf{q}_{1}\mathbf{q}_{0}^{-1})^{\phi{(t)}}\mathbf{q}_{0} = \left ( \cos(\frac{\\gamma}{2}), \sin(\frac{\\gamma }{2})\vec{\mathbf{v}} \right)^{\phi{(t)}}\mathbf{q}_{0}
\label{eq:quat_inter}
\end{equation}
where $\mathbf{q}_{0}^{-1}$ represents the inverse of $\mathbf{q}_{0}$, and the product operation here is the Hamilton product. The $\phi{(t)}$ denotes a monotonically non-decreasing function, ranging from 0 to 1 and controlls the orientation transition from $\mathbf{q}_{0}$ to $\mathbf{q}_{1}$.
When $\phi{(t)}$ equals 0 or 1, $\mathbf{q}_{t}$ will equal to $\mathbf{q}_{0}$ and $\mathbf{q}_{1}$ respectively.
In order to reduce the computational cost brought by the Hamilton product, we further rewrite Eq.~\ref{eq:quat_inter} as a linear combination of the two quaternions $\mathbf{q}_{0}$ and $\mathbf{q}_{1}$ :
\begin{equation}
    \mathbf{q}_{t} = (\mathbf{q}_{1}\mathbf{q}_{0}^{-1})^{\phi{(t)}}\mathbf{q}_{0} = Norm\left[
    \mu(t)\mathbf{q}_{0} + \varepsilon(t) \mathbf{q}_{1} \right]
    \label{eq:liner combination}
\end{equation}
where $Norm[\cdot]$ denotes normalization to ensure the result is a unit quaternion. $\mu(t)$ and $\varepsilon(t)$ are time-dependent coefficients which are determined by $\phi{(t)}$.
One instance of Eq.~\ref{eq:liner combination} is Slerp\cite{shoemake1985animating}, which is a widely used linear quaternion interpolation method with constant rotation speed, assuming $\phi{(t)} = t$:
\begin{equation}
    \mathbf{q}_{t} = (\mathbf{q}_{1}\mathbf{q}_{0}^{-1})^{t}\mathbf{q}_{0} =  \frac{\sin((1-t)\eta)}{\sin(\eta) }\mathbf{q}_{0}+\frac{\sin(t\eta)}{\sin(\\eta)}\mathbf{q}_{1}
    \label{eq:Slerp}
\end{equation}
where $\eta$ is the included angle between $\mathbf{q}_{0}$ and $\mathbf{q}_{1}$ as two vectors, which can be computed by:
\begin{equation}
   \eta = \frac{\mathbf{q}_{1}\cdot \mathbf{q}_{0}}{\left \| \mathbf{q}_{1} \right \|\left \| \mathbf{q}_{0} \right \|}
\end{equation}
where $\cdot$ denotes inner product of two vectors and $\left \| \cdot  \right \|$ is the magnitude of a vector.

\begin{figure}[tb]
\begin{center}
		%\fbox{\rule{0pt}{2in} \rule{0.7\linewidth}{0pt}}
\includegraphics[width=1\linewidth]{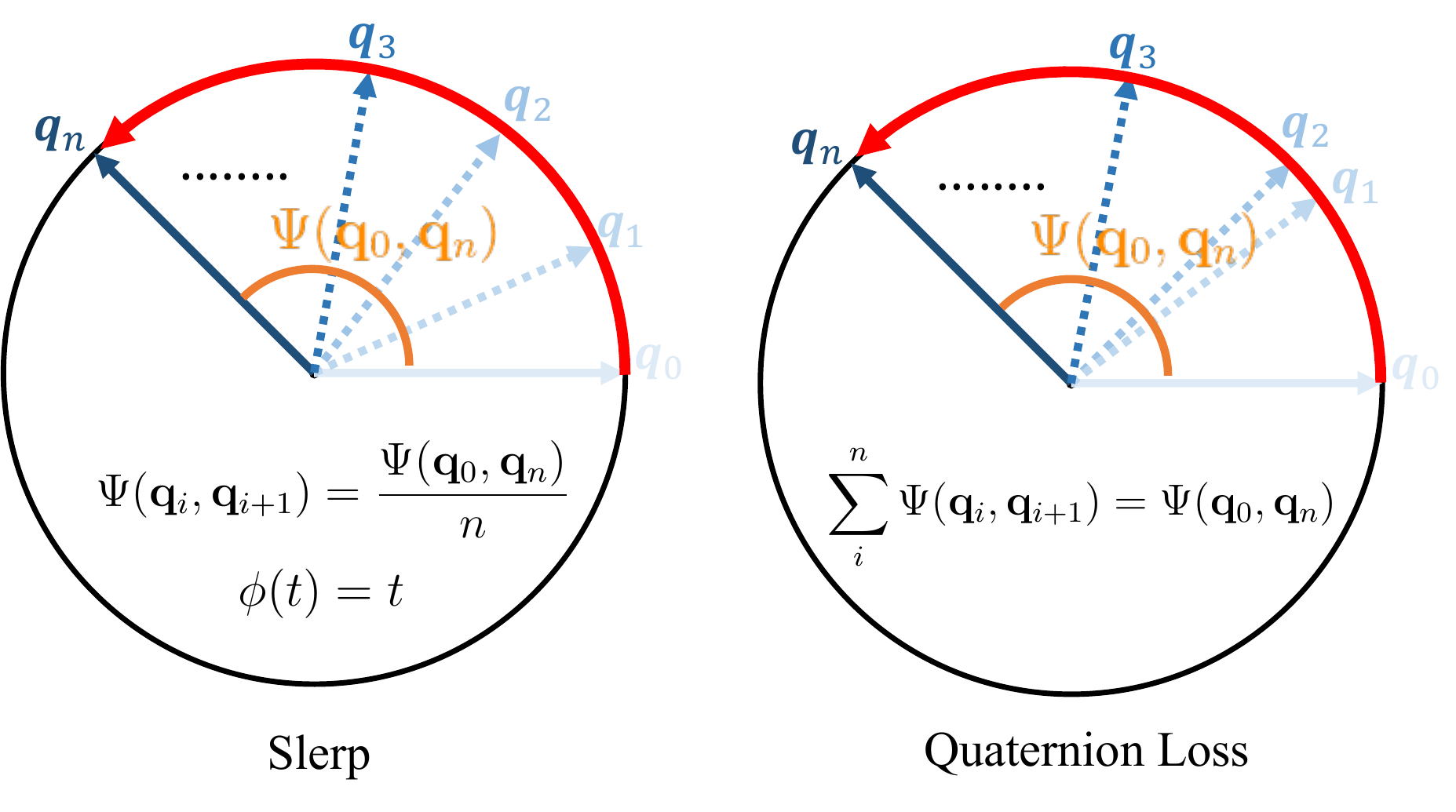}
\end{center}
\caption{Comparison between our quaternion loss and Slerp. The circle represents a 2D projective plane of 4D unit quaternion sphere. The red arch denotes the set of quaternion that satisfies Eq.~\ref{eq:quat_inter}, which ensures smooth orientation transition. The equation in each circle represents the corresponding prior. The remaining symbols can be found in Section~\ref{section:sshrfvq}. As can be seen, both Slerp and quaternion loss has the prior to make sure Eq.~\ref{eq:quat_inter} is satisfied. However Slerp has an additional prior $\phi{(t)} = t$, while our Quaternion loss covers all possible  $\phi{(t)}$, which allows smooth orientation transition at all possible speed.
}
\label{fig:Quat}
%\label{fig:onecol}
\vspace{-0.3cm}
\end{figure}

Instead of generating interpolated poses as psuedo-labels with one specific $\phi(t)$ for supervision, we propose a quaternion loss function to cover all possible joint rotation speeds as following:
\begin{equation}
    E_{quat} = \left \| \sum_{i=1}^{n-1} \Psi(H_{i},H_{i+1}) - \Psi(H_{1},H_{n})  \right \|
    \label{eq:quat}
\end{equation}
where $\Psi$ is the function to compute the rotation angle $\gamma$ between two quaternions, and $H_{i}$ denotes the output hand pose represented in quaternion of frame $i$. In practice, $H_{i}$ is the concatenation of i-th pose vector $\theta_{i}$ and i-th rotation $R_{i}$ to cover all 21 hand joints:
\begin{equation}
    H_{i} = Quaternion(Concatenate[\theta_{i}, R_{i}])
\end{equation}
where $Quaternion$ denotes the transformation from representation of MANO outputs to quaternion representation and $Concatenate$ denotes the concatenation operation. The comparison between the proposed quaternion loss and Slerp is illustrated in Fig.~\ref{fig:Quat}. 

To understand the proposed quaternion loss, two points are important. One is that the quaternion interpolation is essentially finding a rotation curve through a fixed rotation axis between two poses, as suggested by Eq.~\ref{eq:quat_inter}. The other is that the rotation angle $\gamma$ in the quaternion space relates to included angle $\eta$ in vector space, as indicated by Eq.~\ref{eq:liner combination}. Specifically:
\begin{equation}
    \cos (\frac{\gamma }{2}) =  \cos(\eta)
\label{eq:quat_angle}
\end{equation}
which provides an efficient way to compute $\gamma$ and is derived from:
\begin{equation}
    \Re \left ( \mathbf{q}_{1}\mathbf{q}_{0}^{-1} \right ) = \Re\left ( \cos(\frac{\gamma}{2}), \sin(\frac{\gamma }{2})\vec{\mathbf{v}} \right) = \left \| \mathbf{q}_{1} \right \|\left \| \mathbf{q}_{0} \right \|\cos( \eta)
\label{eq:quat_math}
\end{equation}
where $\Re$ represents the real part of a quaternion, $\mathbf{q}_{0}$, $\mathbf{q}_{1}$, $\gamma$ and $\vec{\mathbf{v}}$ are the same as before. $\left \| \cdot  \right \|$ is the magnitude of a vector. $\eta$ denotes the included angle between $\mathbf{q}_{0}$ and $\mathbf{q}_{1}$ as two vectors. Eq.~\ref{eq:quat_math} can be deduced by comparing the inner product and the Hamilton product of the $\mathbf{q}_{0}$ and $\mathbf{q}_{1}$.

%\subsubsection{T\&S  consistency loss}
\subsubsection{T\&S  Consistency Regularization}
\label{CT&S}
We introduce a regularization term on texture and shape to consider consistency of hand appearance in videos. Since texture is coupled with light, our T\&S loss is formulated as:
\begin{equation}
    E_{T\&S} = \sum_{i=1}^{n} \left \| C_{i}^{L} - \overline{C^{L}} \right \| +\sum_{i=1}^{n} \left \| \beta_{i} - \overline{\beta} \right \|
\end{equation}
where $C_{i}^{L}$ and $\beta_{i}$ are the i-th lighted texture and shape of the sequential reconstruction output, $\overline{C_{i}^{L}}$ and $\overline{\beta_{i}}$ are the corresponding average of the sequential output. The lighted texture $C_{i}^{L}$ is computed following \cite{kato2018neural}: 
\begin{equation}
    C_{i}^{L} = (l^{a}l_{c}^{a}+(n^{d}\cdot n_{i})l^{d}l_{c}^{d})C_{i}
    \label{eq:lighted_texture}
\end{equation}
where $n_{i}$ is the normal direction of $C_{i}$ in canonical zero pose \cite{MANO:SIGGRAPHASIA:2017} and the rest are defined in Section~\ref{sec:encoding}.

A low standard deviation of sequential hand appearance reconstruction from video sequences is promoted by this loss function.
\subsubsection{Training Objective}
Our overall training loss $E_{S^{2}HAND(V)}$ consists of three parts, including a ${\rm {S}^{2}HAND}$ loss $E_{S^{2}HAND}$, a quaternion loss $E_{quat}$ and a $T\&S$ loss $E_{T\&S}$:
\begin{equation}
E_{S^{2}HAND(V)}= E_{S^{2}HAND} + w_{quat}E_{quat} + w_{ts}E_{T\&S}
\end{equation}
where $E_{S^{2}HAND}$ is the same as that in Section~\ref{image TO}. For $E_{quat}$ and $E_{T\&S}$, please refer to Section~\ref{Sql} and Section~\ref{CT&S} respectively. The constant weights $w_{quat}$ and $w_{ts}$ are used to balance the three terms.

\section{Experiments}
\subsection{Datasets}
\label{Datasets}
We evaluate the proposed methods on three datasets. Two of them (FreiHAND and HO-3D) are challenging realistic datasets, aiming for assessing 3D joints and 3D meshes with hand-object interaction. The results are reported through the online submission systems \footnote{\scriptsize {https://competitions.codalab.org/competitions/21238}}${}^{,}$\footnote{\scriptsize {https://competitions.codalab.org/competitions/22485}}. The remaining one (STB) is a hand-only video dataset. Besides, we adopt another dataset (YT 3D) to provide in-the-wild data.

The FreiHAND dataset \cite{zimmermann2019freihand} is a large-scale real-world dataset,
which contains 32,560 training samples and 3,960 test samples. For each training sample, one real RGB image and extra three images with different synthetic backgrounds are provided.
Part of the sample is a hand grabbing an object, but it does not provide any annotations for the foreground object, which poses additional challenges.

The HO-3D dataset \cite{hampali2020honnotate} collects color images of a hand interacting with an object.
The dataset is made of 68 sequences, totaling 77,558 frames of 10 users manipulating one among 10 different objects.
The training set contains 66,034 images and the test set contains 11,524 images.
The objects in this dataset are larger than that in FreiHAND, thus resulting in larger occlusions to hands.
We use this dataset in two cases. In the case of self-supervised hand reconstruction from image collections with ${\rm {S}^{2}HAND}$, we do not use the sequence information provided by HO-3D and mix all sequences as a image collection. In the case of self-supervised hand reconstruction from video sequences with ${\rm {S}^{2}HAND(V)}$, we make use of the sequence information and
compose the training batch accordingly. Details can be found in Section~\ref{implementation}.

The STB dataset \cite{zhang2017hand} is a hand-only dataset, which contains 12 sequences with 18000 frames in total. RGB images along with depth-images, 2D and 3D joint annotations are provided. We follow the splits in \cite{zhang2017hand}, using 10 sequences for training and 2 sequences for evaluation. We select this dataset to validate the proposed methods in the hand-only scenario.

The YT 3D dataset \cite{kulon2020weakly} contains 116 in-the-wild videos, which is comprised of 102 train videos, 7 validation videos and 7 test videos, along with 47125, 1525 and 1525 hand annotations.
We only use this dataset as extra in-the-wild training data in Section~\ref{HR in the wild} without any annotations. Since the 102 train videos are edited with lots of cutaway and only part of videos are accessible due to copyright issues, we preprocess this dataset by filtering out unavailable videos and discontinuous hand motion frames according to the detected 2D keypoints, yielding 34 train videos containing 21628 frames with detected 2D keypoints.

\subsection{Evaluation Metrics}
\label{evaluation metrics}
We evaluate 3D hand reconstruction by evaluating 3D joints and 3D meshes. For 3D joints, we report the \textbf{mean per joint position error} (MPJPE) in the Euclidean space for all joints on all test frames in \textit{cm} and the \textbf{area under the curve} (AUC) of the PCK $\rm AUC_{J}$.
Here, the PCK refers to the percentage of correct keypoints.
For 3D meshes, we report the \textbf{mean per vertex position error} (MPVPE) in the Euclidean space for all joints on all test frames in \textit{cm} and the AUC of the percentage of correct vertex $\rm AUC_{V}$.
We also compare the F-score \cite{knapitsch2017tanks} which is the harmonic mean of recall and precision for a given distance threshold. We report distance threshold at 5\textit{mm} and 15\textit{mm} and report F-score of mesh vertices at 5\textit{mm} and 15\textit{mm} by ${\rm F}_5$ and ${\rm F}_{15}$.
Following the previous works \cite{hampali2020honnotate,zimmermann2019freihand}, we compare aligned prediction results with Procrustes alignment, and all 3D results are evaluated by the online evaluation system on FreiHAND and HO-3D. 
For 2D joints, we report the MPJPE in \textit{pixel} and the curve plot of \textbf{fraction of joints within distance}.
For smooth hand reconstruction, we report the \textbf{acceleration error} (ACC-ERR) and the \textbf{acceleratio}n(ACC) which are first proposed in \cite{kanazawa2019learning}. ACC-ERR measures average difference between ground truth acceleration and the acceleration of the predicted 3D joints in $mm/s^{2}$ while ACC calculates mean acceleration of the predicted 3D joints in $mm/s^{2}$.
Generally, lower ACC-ERR and ACC indicate smoother sequence predictions.
For shape and texture consistency in sequence predictions, we report corresponding \textbf{standard deviations (S.D.)}, in which low deviation means coherent and consistent sequence predictions. Specifically, we compute texture S.D. and shape S.D., which are the average of per dimensional S.D. of the lighted textures and shape parameters in sequence predictions respectively.

\begin{table*}[t]
\begin{minipage}{1\linewidth}
\centering
\small
{\def\arraystretch{1} \tabcolsep=1.7em 
\caption{Comparison of main results on the FreiHAND testing set. The performance of our self-supervised method ${\rm {S}^{2}HAND}$ is comparable to the recent fully-supervised and weakly-supervised methods. \cite{spurr2020weakly}* also uses the synthetic training data with 3D supervision}. Note that FreiHAND is not presented with video sequences, which disables learning of ${\rm {S}^{2}HAND(V)}$.
\label{table:sc-ssl}
% \scalebox{1}{
\begin{tabular}{cc|cc|cccc}
        \hline
        Supervision & Method & $\rm {AUC}_{J}$$\uparrow$ & MPJPE$\downarrow$&$\rm {AUC}_{V}$$\uparrow$ &  MPVPE$\downarrow$ & ${\rm F}_5$$\uparrow$ & ${\rm F}_{15}$$\uparrow$\\
        \hline
        \multirow{4}{*}{3D} &\cite{zimmermann2019freihand}(2019)& 0.35 & 3.50 & 0.74 & 1.32 & 0.43 & 0.90  \\%Zimmermann 0.427 & 0.895
        &\cite{hasson2019learning}(2019) & 0.74 & 1.33 & 0.74 & 1.33 & 0.43 & 0.91\\%Hasson 0.429 & 0.907
        &\cite{boukhayma20193d}(2019) & \textbf{0.78} & \textbf{1.10} & \textbf{0.78} & \textbf{1.09} & \textbf{0.52} & \textbf{0.93}\\%Boukhayma  0.516 & 0.934
        %& - & 7.6 7.4 0.681 0.973&\\&&\\
        &\cite{qian2020parametric}(2020) & \textbf{0.78} & \underline{1.11} & \textbf{0.78}& \underline{1.10} & \underline{0.51} & \textbf{0.93}\\%Qian &0.508&0.930
        %& Moon (2020) \cite{moon2020i2l} & - & 0.74 & - & 0.76  & 0.681 & 0.973\\
        %\cdashline{2-8}
        %& Ours &\\
        \hline
        \multirow{1}{*}{2D} &\cite{spurr2020weakly}(2020)* & \textbf{0.78} & 1.13 & - & - & - & -\\%Spurr
        %\hdashline
        %\cdashline{2-8}
        %& Ours &\\
        \hline
        - & ${\rm {S}^{2}HAND}$ & \underline{0.77} & 1.18 & \underline{0.77} & 1.19 & 0.48 & \underline{0.92}\\%0.483 & 0.917
\hline
\end{tabular}
% }
}
\end{minipage}

\vspace{0.2cm}

\par
\begin{minipage}{\linewidth}
\centering
\small
{\def\arraystretch{1} \tabcolsep=1.65em 
\caption{Comparison of main results on the HO-3D testing set. Without using any object information and hand annotation, our ${\rm {S}^{2}HAND}$ model performs comparable with the recent fully-supervised methods \cite{hasson2020leveraging}. Further with the temporal constraints, our ${\rm {S}^{2}HAND(V)}$ even surpasses \cite{hasson2020leveraging}.}

\label{table:sc-ssl-ho}
\begin{tabular}{cc|cc|cccc}
        \hline
        Supervision & Method & $\rm {AUC}_{J}$$\uparrow$ & MPJPE$\downarrow$& $\rm {AUC}_{V}$$\uparrow$ & MPVPE$\downarrow$ & ${\rm F}_5$$\uparrow$ & ${\rm F}_{15}$$\uparrow$\\
        \hline
        \multirow{3}{*}{3D}
        &\cite{hasson2019learning}(2019) & - & - & - & 1.30 & 0.42 & 0.90\\
        &\cite{hampali2020honnotate}(2020) & - & - & - & \textbf{1.06} & \textbf{0.51} & \textbf{0.94} \\
        &\cite{hasson2020leveraging}(2020) & \underline{0.773} & \underline{1.11} & 0.773 & 1.14 & 0.43 & \underline{0.93} \\
        \hline
         \multirow{2}{*}{-}  & ${\rm {S}^{2}HAND}$ & \underline{0.773} & 1.14 & \underline{0.777} & 1.12 & 0.45 & \underline{0.93}\\
         & ${\rm {S}^{2}HAND(V)}$ & \textbf{0.780} & \textbf{1.10} & \textbf{0.781} & \underline{1.09} & \underline{0.46} & \textbf{0.94}\\
        \hline
\end{tabular}
}
\end{minipage}

\vspace{0.2cm}
\par

\begin{minipage}{\linewidth}
\centering
\small
{\def\arraystretch{1} \tabcolsep=1.05em 
\caption{Ablation studies on different losses used in our method on the {FreiHAND} testing set. Refer to Section~\ref{HR from images} for details.}
\label{table:ablation}
\begin{tabular}{ccccc|cc|cccc}
    \hline
    \multicolumn{5}{c|}{Losses} &
    \multirow{2}{*}{MPJPE$\downarrow$}&
    \multirow{2}{*}{MPVPE$\downarrow$} &
    \multirow{2}{*}{$\rm {AUC}_{J}\uparrow$} &  \multirow{2}{*}{$\rm {AUC}_{V}\uparrow$} & \multirow{2}{*}{${\rm F}_5\uparrow$} & \multirow{2}{*}{${\rm F}_{15}\uparrow$} \\
    \cline{1-5}
    ${E}_{loc}$ & ${E}_{regu}$ & ${E}_{ori}$ & ${E}_{2d}$, ${E}_{con}$ & ${E}_{photo}$&&&&&&\\
    \hline
    $\checkmark$ & & & & & 1.97 & 2.31 & 0.611 & 0.545 & 0.257 & 0.763 \\%116
    $\checkmark$ &$\checkmark$ & & & & 1.54 & 1.58 & 0.696 & 0.687 & 0.387 & 0.852 \\%116
    $\checkmark$ &$\checkmark$ & $\checkmark$ & & & 1.24 & 1.26 & 0.754 & 0.750 & 0.457 & 0.903\\%82
    $\checkmark$ &$\checkmark$ & $\checkmark$ & & $\checkmark$ & 1.22 & 1.24 & 0.759 & 0.754 & 0.468 & 0.909\\%86
    $\checkmark$ &$\checkmark$ & $\checkmark$ & $\checkmark$ & ~ & 1.19& 1.20 & 0.764 & 0.763 & 0.479 & 0.915\\%83
    $\checkmark$ &$\checkmark$ & $\checkmark$ & $\checkmark$ & $\checkmark$ & \textbf{1.18} & \textbf{1.19} & \textbf{0.766} & \textbf{0.765} & \textbf{0.483} & \textbf{0.917}\\%86
   
    \hline
    \end{tabular}
}
\end{minipage}
\vspace{-0.5cm}
\end{table*}

\subsection{Implementation Details}
\label{implementation}
Pytorch \cite{paszke2017automatic} is used for implementation. For the 3D reconstruction network, the EfficientNet-b0 \cite{tan2019efficientnet} is pre-trained on the ImageNet dataset. The 2D keypoint estimator along with the 2D-3D consistency loss is optionally used. If we train the whole network with the 2D keypoint estimator, a stage-wise training scheme is used.
We train the 2D keypoint estimator and 3D reconstruction network by 90 epochs separately, where $E_{3d}$ and $E_{2d}$ are used, respectively. The initial learning rate is ${10}^{-3}$ and reduced by a factor of 2 after every 30 epochs. We perform data augmentations (including cropping) on input images following \cite{hasson2019learning}, where the projection matrix is modified accordingly to ensure correct rendering.

\begin{figure}[tb]
\vspace{-0.32cm}
    \setlength{\abovecaptionskip}{0.cm}
	%\begin{center}
	%\fbox{\rule{0pt}{2in} \rule{0.7\linewidth}{0pt}}
	\includegraphics[width=1\linewidth]{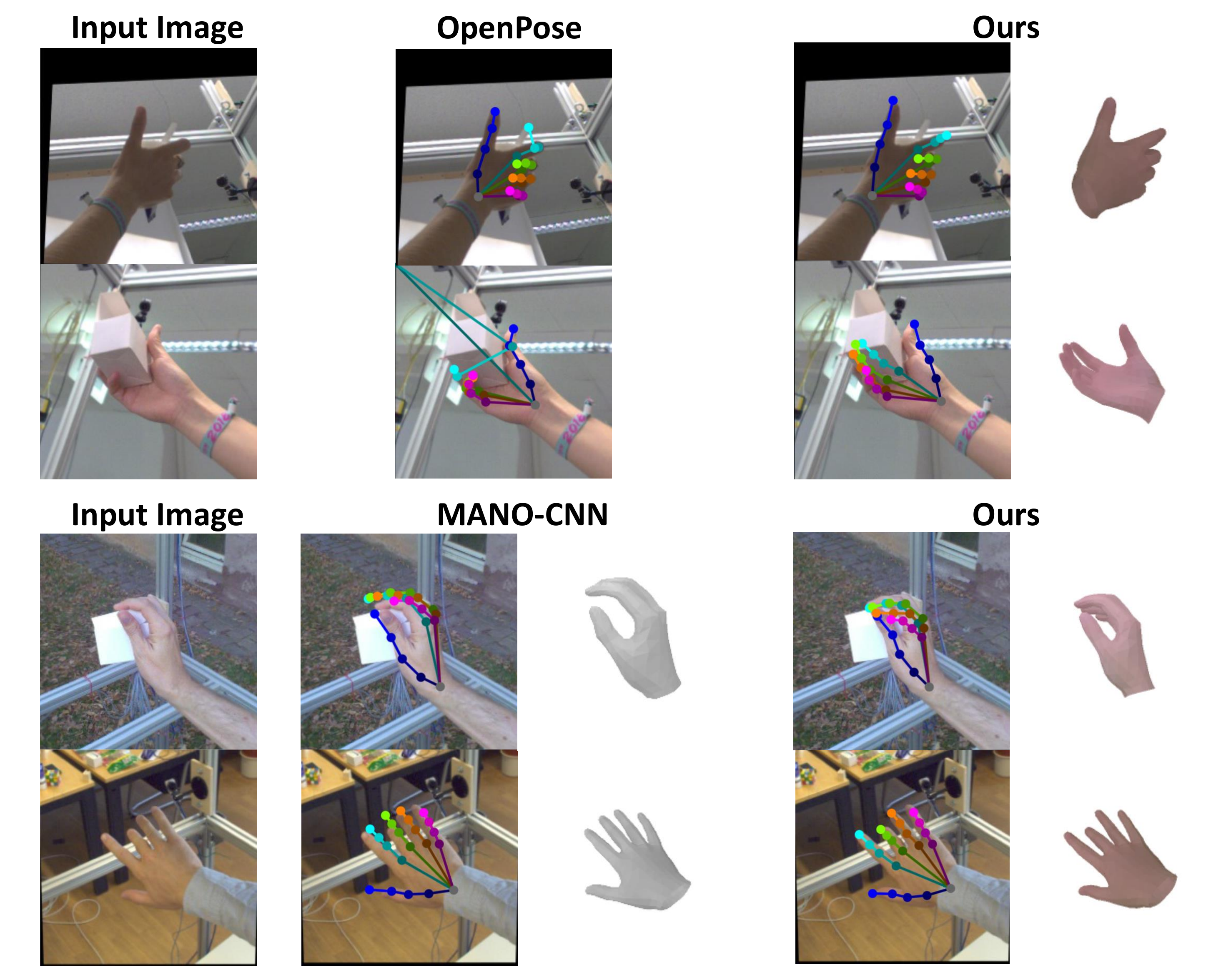}
	%\end{center}
	\caption{Qualitative comparison to OpenPose \cite{cao2019openpose} and MANO-CNN on the FreiHAND testing set. For OpenPose, we visualize the detected 2D keypoints. For our method and MANO-CNN, we visualize both the projected 2D keypoints and 3D mesh}.
	\label{fig:qualitative}
	%\label{fig:onecol}
\vspace{-0.8cm}
\end{figure}

When training ${\rm {S}^{2}HAND}$ with image collections, we finetune the whole network with $E_{S^{2}HAND}$ by 60 epochs with the learning rate initialized to $2.5\times {10}^{-4}$, and reduced by a factor of 3 after every 20 epochs. We use Adam \cite{kingma2014adam} to optimize the network weights with a batch size of 64. We train our model on two NVIDIA Tesla V100 GPUs, which takes around 36 hours for training on FreiHAND.
Otherwise, when training ${\rm {S}^{2}HAND(V)}$ with $E_{S^{2}HAND(V)}$, the input remains 4D (B,C,H,W), but the sampling strategy is different. Specifically, for a training batch, we first randomly sample $m$ sequences from the training sequences, then randomly sample $n$ frames in each of the sampled sequences, finally combine these frames to compose a batch. This results in a batch size of $mn$. In our experiments, we set $m$ equal to $64 // n$, where 64 follows the batch
size in training ${\rm {S}^{2}HAND}$ and $//$ represents floor division. We find $n$ equals 3 (see Section~\ref{HR from sq} for the ablation study) gets the best performance, thus we use it as the default setting. Notice that all the sampling procedures take place during training. For the stability of this sampling strategy, please refer to Section~2 of the supplementary materials.
 The learning rate, the reducing schedule, and the Adam optimizer are set the same as before. 
For the weighting factors, we set \mbox{$w_{3d}=1$}, $w_{2d}=0.001$, {$w_{cons}$}$=0.0002$, $w_{geo}=0.001$, $w_{photo}=0.005$, $w_{quat}=0.05$, $w_{ts}=0.01$, $w_{regu}=0.01$, $w_{ori}=100$, $w_{SSIM}=0.2$, $w_{C}=0.5$, $w_{s}=10$ and $w_{J}=10$. {Here the weighting factors need to regard different fundamental units of the losses, thus the magnitudes do not strictly imply the importance. 
For how to select these weights and the weights sensitivity, please refer to the supplementary materials.}

\subsection{Comparison with State-of-the-art Methods}
We give comparison on FreiHAND with four recent model-based fully-supervised methods \cite{boukhayma20193d,hasson2019learning, qian2020parametric,zimmermann2019freihand} and a state-of-the-art weakly-supervised method \cite{spurr2020weakly} in \mbox{Table~\ref{table:sc-ssl}}.
Note that \cite{moon2020i2l} is not included here since it designs an advanced ``image-to-lixel" prediction instead of directly regressing MANO parameters.
Our approach ${\rm {S}^{2}HAND}$ focuses on providing a self-supervised framework with lightweight components, where the hand regression scheme is still affected by highly non-linear mapping.
Therefore, we make a fairer comparison with popular model-based methods \cite{boukhayma20193d,hasson2019learning,qian2020parametric,zimmermann2019freihand} to demonstrate the performance of this self-supervised approach.
Without using any annotation, our approach ${\rm {S}^{2}HAND}$ outperforms \cite{hasson2019learning,zimmermann2019freihand} on all evaluation metrics and achieves comparable performance to \cite{boukhayma20193d,qian2020parametric}.
\cite{spurr2020weakly} only outputs 3D pose, and its pose performance is slightly better than our results on FreiHAND test set but with much more training data used including RHD dataset \cite{zimmermann2017learning} (with 40,000+ synthetic images and 3D annotations) as well as 2D ground truth annotation of the FreiHAND.

In the hand-object interaction scenario, we compare with three recent fully-supervised methods on HO-3D in Table~\ref{table:sc-ssl-ho}.
Compared to the hand branch of \cite{hasson2019learning}, both of our self-supervised models ${\rm {S}^{2}HAND}$ and ${\rm {S}^{2}HAND(V)}$ show higher mesh reconstruction performance where we get a 14\% and 16\% reduction in MPVPE respectively.
Compared with \cite{hasson2020leveraging}, which is a fully-supervised joint hand-object pose estimation method, our ${\rm {S}^{2}HAND}$ obtains comparable joints and shape estimation results, while our ${\rm {S}^{2}HAND(V)}$ even surpasses \cite{hasson2020leveraging}.
\cite{hampali2020honnotate} gets slightly better shape estimation results than ours, probably due to the fact that they first estimate 2D keypoint positions using heatmaps and then fit MANO model to these keypoints.
Since \cite{spurr2020weakly} utilizes a totally different version of HO-3D that is published with HANDS 2019 Challenge\footnote{\scriptsize{https://sites.google.com/view/hands2019/challenge}}, we do not compare with it on HO-3D (not shown in Table~\ref{table:sc-ssl-ho}) as we do on FreiHAND.

\begin{figure}[t]
\vspace{-0.2cm}
    \centering
	\includegraphics[width=1\linewidth]{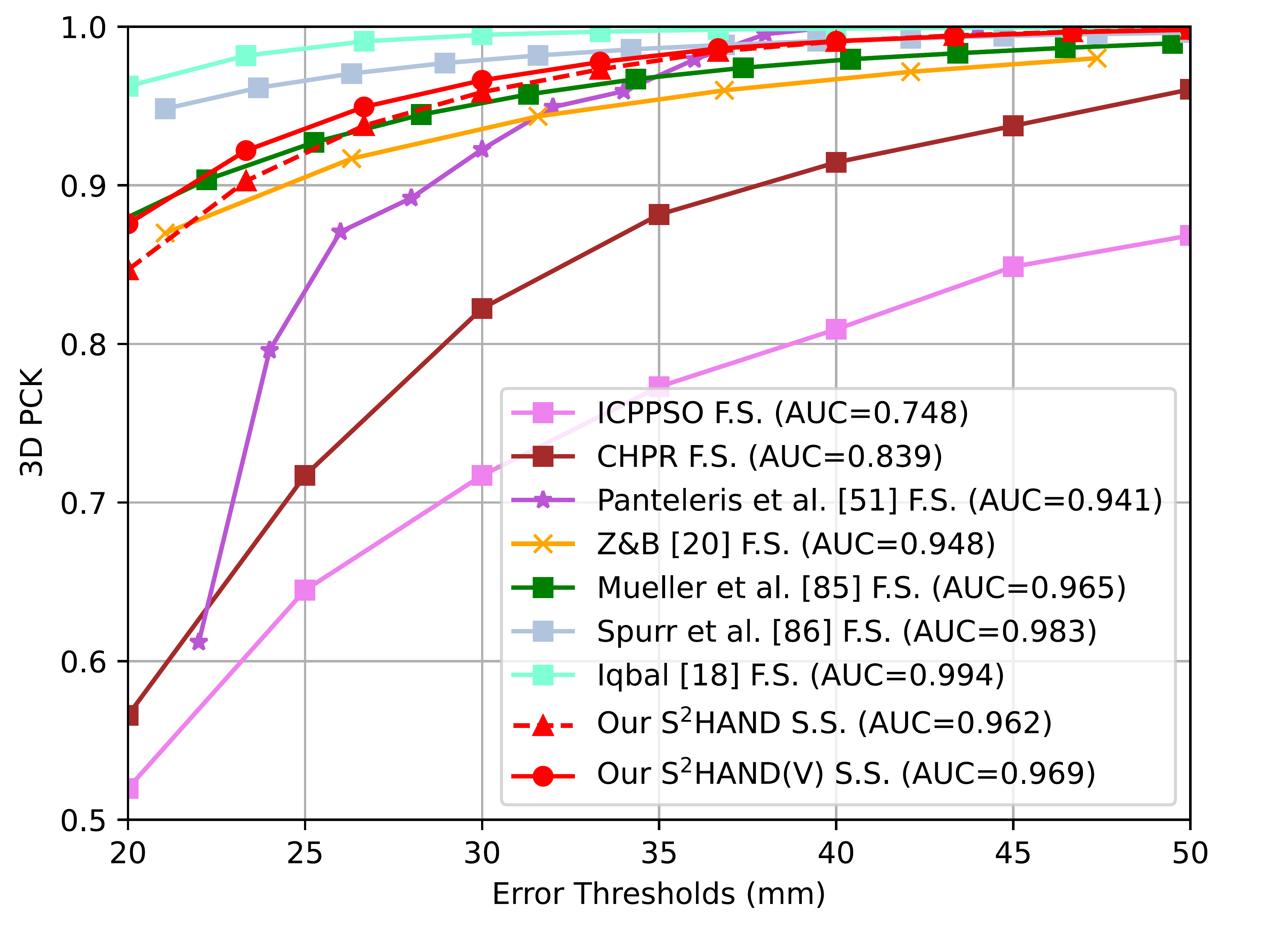}
    \caption{Comparisons between our proposed self-supervised (S.S.) methods and other SOTA fully-supervised (F.S.) methods on the STB dataset (hand-only scenario).}
	\label{fig:stb}
\vspace{-0.6cm}
\end{figure}

\begin{table*}[t]
\begin{minipage}{\linewidth}
\centering
\small
{\def\arraystretch{1} \tabcolsep=1.2em 
\caption{Comparison of the accuracy of different motion-related constraints on the HO-3D testing set. Quaternion loss shows its effectiveness over the similar loss functions on modeling smoothness.}
\label{table:Quat comparison}
\begin{tabular}{c|cc|cccc}
        \hline
        %Supervisions & Pose \quad\quad & Mesh \quad & F-score & F-score\\
        %Terms & Error \downarrow & Error \downarrow & @5mm \uparrow & @15mm \uparrow\\
        Method & $\rm {AUC}_{J}$$\uparrow$ & MPJPE$\downarrow$& $\rm {AUC}_{V}$$\uparrow$ & MPVPE$\downarrow$ & ${\rm F}_5$$\uparrow$ & ${\rm F}_{15}$$\uparrow$\\%& STA MPJPE* $\downarrow$
        \hline

        ${\rm {S}^{2}HAND}$ & 0.773 & 1.14 & 0.777 & 1.12 & 0.45 & 0.93\\%2 (3D branch) & 4.81 \textcolor{red}{to-update}
        % & Ours & 4.73 & 0.772 & 1.14 & 0.775 & 1.13 & 0.45 & 0.93 \\%7
        % & Ours (w/ ref.)\\
        ${\rm {S}^{2}HAND(V)}$ (w/ Temporal Loss\cite{yang2020seqhand}) & 0.771 & 1.15  & 0.773 & 1.14 & 0.44 & 0.93\\
        ${\rm {S}^{2}HAND(V)}$ (w/ Smooth Loss\cite{liu2021semi}) & 0.774 & 1.13 & 0.776 & 1.12 & 0.45 & 0.93\\
        ${\rm {S}^{2}HAND(V)}$ (w/ Quaternion Loss) & 0.779 &\textbf{1.10} & \textbf{0.781} & \textbf{1.09} & \textbf{0.46} & \textbf{0.94}\\
        %- & Ours (w/ RF) & - & - & - & - & - & - &\\
        \hline
        ${\rm {S}^{2}HAND(V)}$ (w/ Quaternion Loss, T\&S Loss) & \textbf{0.780}  & \textbf{1.10}  & \textbf{0.781} & \textbf{1.09} & \textbf{0.46} & \textbf{0.94}\\
        \hline
\end{tabular}
}
\end{minipage}

\vspace{0.2cm}
\par

\begin{minipage}{\linewidth}
\centering
\small
{\def\arraystretch{1} \tabcolsep=1.1em 
\caption{Comparison of the smoothoness performance of different motion-related constraints on the HO-3D dataset. Quaternion loss gives the smoothest predictions and is highly in line with ACC and ACC-ERR.}
\label{table:smooth comparison}
\begin{tabular}{c|ccc|cc}
        \hline
        \multirow{2}{*}{Method}& \multicolumn{3}{c|}{Train set} & \multicolumn{2}{c}{Test set}\\
        \cline{2-6}
        & ACC-ERR$\downarrow$ &	ACC$\downarrow$ & Quaternion Loss$\downarrow$ & ACC$\downarrow$ & Quaternion Loss$\downarrow$\\
        \hline
        ${\rm {S}^{2}HAND}$ & 2.85 &	2.29 &	0.014 &	3.68 &	0.020\\
        ${\rm {S}^{2}HAND(V)}$ (w/ Temporal Loss\cite{yang2020seqhand})&2.57 & 1.96 & 0.011 & 2.87 & 0.015\\
        ${\rm {S}^{2}HAND(V)}$ (w/ Smooth Loss\cite{liu2021semi}) & 2.89 & 2.29 & 0.014 & 2.89 & 0.015\\
        ${\rm {S}^{2}HAND(V)}$ (w/ Quaternion Loss) & \textbf{2.23} & \textbf{1.59} & \textbf{0.008} & \textbf{2.81} & \textbf{0.014}\\
        \hline
\end{tabular}
}
\end{minipage}
\vspace{-0.5cm}
\end{table*}

{In the hand-only scenario, we compare with the fully-supervised methods \cite{panteleris2018using, zimmermann2017learning, mueller2018ganerated, spurr2018cross, iqbal2018hand} on STB in Fig.~\ref{fig:stb}. Some state-of-the-art fully-supervised methods (e.g. \cite{baek2020weakly} with 0.995 AUC and \cite{ theodoridis2020cross} with 0.997 AUC) are not included in Fig.~\ref{fig:stb}, because their PCK values are not provided and they have similar performance with \cite{iqbal2018hand}. The proposed self-supervised method ${\rm {S}^{2}HAND}$ outperforms some previous fully-supervised methods \cite{panteleris2018using, zimmermann2017learning, mueller2018ganerated}, and ${\rm {S}^{2}HAND(V)}$ further improves the performance of ${\rm {S}^{2}HAND}$. On the other side, there is a certain gap between ours and the recent fully-supervised methods \cite{spurr2018cross, iqbal2018hand} since they have direct 3D supervision.}

In Fig.~\ref{fig:qualitative}, we show 2D keypoint detection from OpenPose~\cite{cao2019openpose} and our ${\rm {S}^{2}HAND}$ results of difficult samples. We also compare the reconstruction results with MANO-CNN, which directly estimates MANO parameters with a CNN \cite{zimmermann2019freihand}, but we modify its backbone to be the same as ours. Our results are more accurate and additionally with texture.

\begin{figure}[t]
    \centering
	\includegraphics[width=0.99\linewidth]{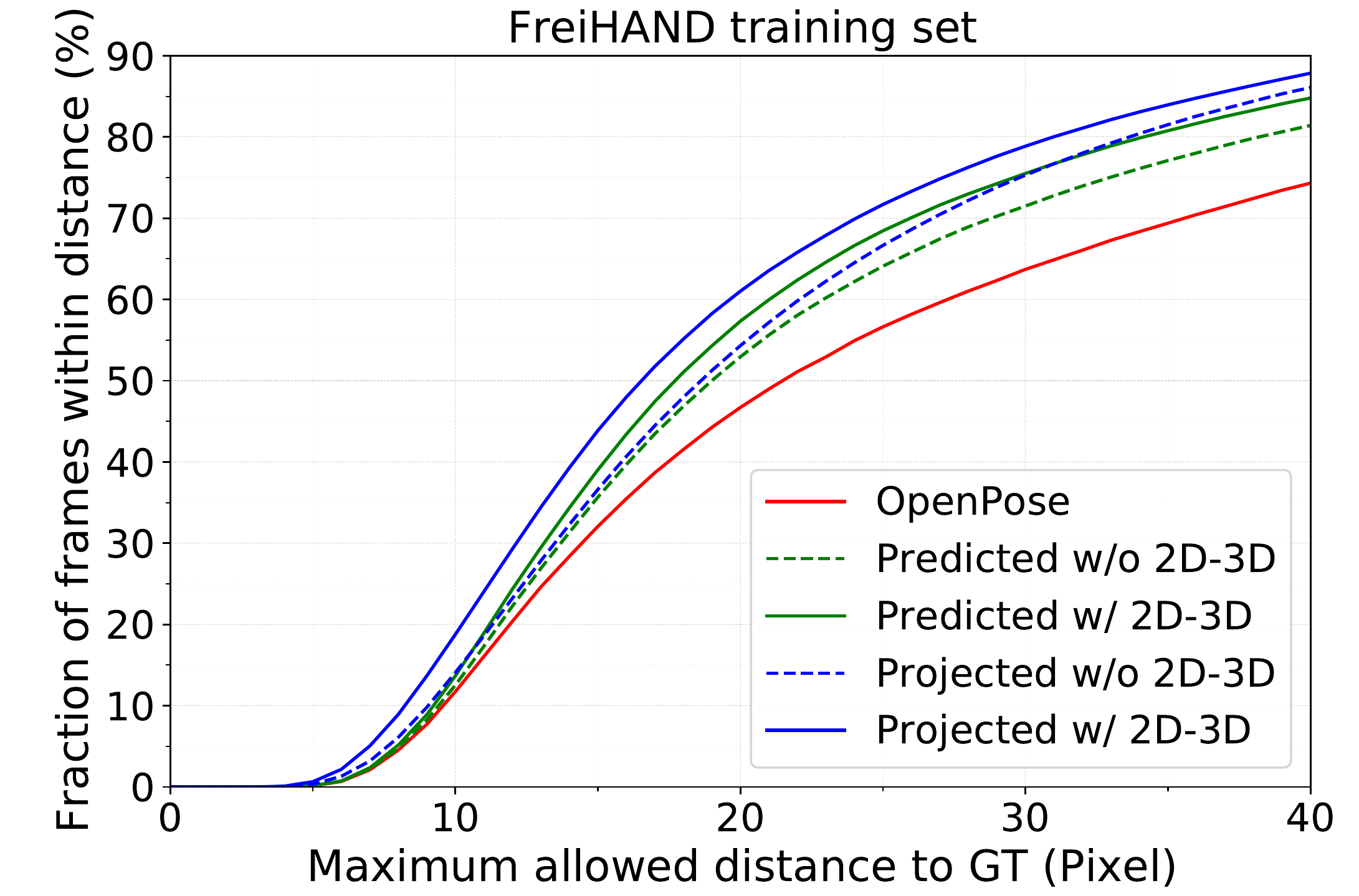}
    \caption{A comparison of 2D keypoint sets used or outputted at the training stage on FreiHAND. The fraction of frames within the maximum joint distance is plotted. Refer to \mbox{Section~\ref{HR from images}} for details.}
	\label{fig:2d}
\vspace{-0.5cm}
\end{figure}

\subsection{Self-comparison}
In Section~\ref{HR from images} and Section
~\ref{HR from sq}, we conduct extensive self-comparisons to verify the effectiveness of each component of our models ${\rm {S}^{2}HAND}$ and ${\rm {S}^{2}HAND(V)}$. In Section~\ref{HR in the wild}, we explore the effect of absorbing extra in-the-wild video sequences with ${\rm {S}^{2}HAND(V)}$.

\subsubsection{Hand Reconstruction from Image Collections}
\label{HR from images}
For self-supervised hand reconstruction from image collection with ${\rm {S}^{2}HAND}$, we conduct ablation studies on FreiHAND, since it is a widely used challenging dataset for hand pose and shape estimation from single RGB images.

First, we give evaluation results on \mbox{FreiHAND} of settings with different components along with corresponding loss terms used in the network in Table~\ref{table:ablation}.
The baseline only uses the 3D branch with $E_{loc}$ and $E_{regu}$, then we add $E_{ori}$ which helps the MPJPE and MPVPE decrease by 19.5\%.
After adding the 2D branch with $E_{2d}$ and the 2D-3D consistency constrain $E_{cons}$, the MPJPE and MPVPE further reduce by 4\%.
The $E_{photo}$ slightly improves the pose and shape estimation results.

\begin{table}[t]
\begin{minipage}{\linewidth}
\centering
\small
{\def\arraystretch{1} \tabcolsep=0.75em 
\caption{Comparison of self-supervised results and weakly-supervised results. Refer to Section~\ref{HR from images} for details.}
\label{table:ssl-wsl}
\begin{tabular}{c|c|cccccc}
    \hline
    Dataset & Method & $\rm AUC_{J}$$\uparrow$ & $\rm AUC_{V}$$\uparrow$ & $\rm F_5$$\uparrow$ & $\rm F_{15}$$\uparrow$\\%PA MPJPE & PA MPVPE
    \hline
    \multirow{2}{*}{FreiHAND} & WSL & 0.730 & 0.725 & 0.42 & 0.89 \\%132   & 1.37 & 1.39
    %&WSL(2D-3D)&\\%
    %&SSL(3D) & \\%
    &SSL & \textbf{0.766} & \textbf{0.765} & \textbf{0.48} & \textbf{0.92} \\%86 & 1.18 & 1.19
    \hline
    \multirow{2}{*}{HO-3D} & WSL & 0.765 & 0.769  & 0.44 & \textbf{0.93}  \\%3 (3D) & 1.18 & 1.15
    %&WSL(2D-3D)\textcolor{red}{to-update} & 1.20 & 1.18 & 0.43 & 0.93\\%
    &SSL & \textbf{0.773} & \textbf{0.777}  & \textbf{0.45} & \textbf{0.93} \\%2 (3D) & 1.14 & 1.12
    %&SSL & 1.14 & 1.13 & 0.45 & 0.93\\%7 (2D-3D)
    \hline
    \end{tabular}
}
\end{minipage}
\vspace{-0.6cm}
\end{table}

\begin{figure*}[t]
\vspace{-0.25cm}
\centering

\includegraphics[width=0.95\linewidth]{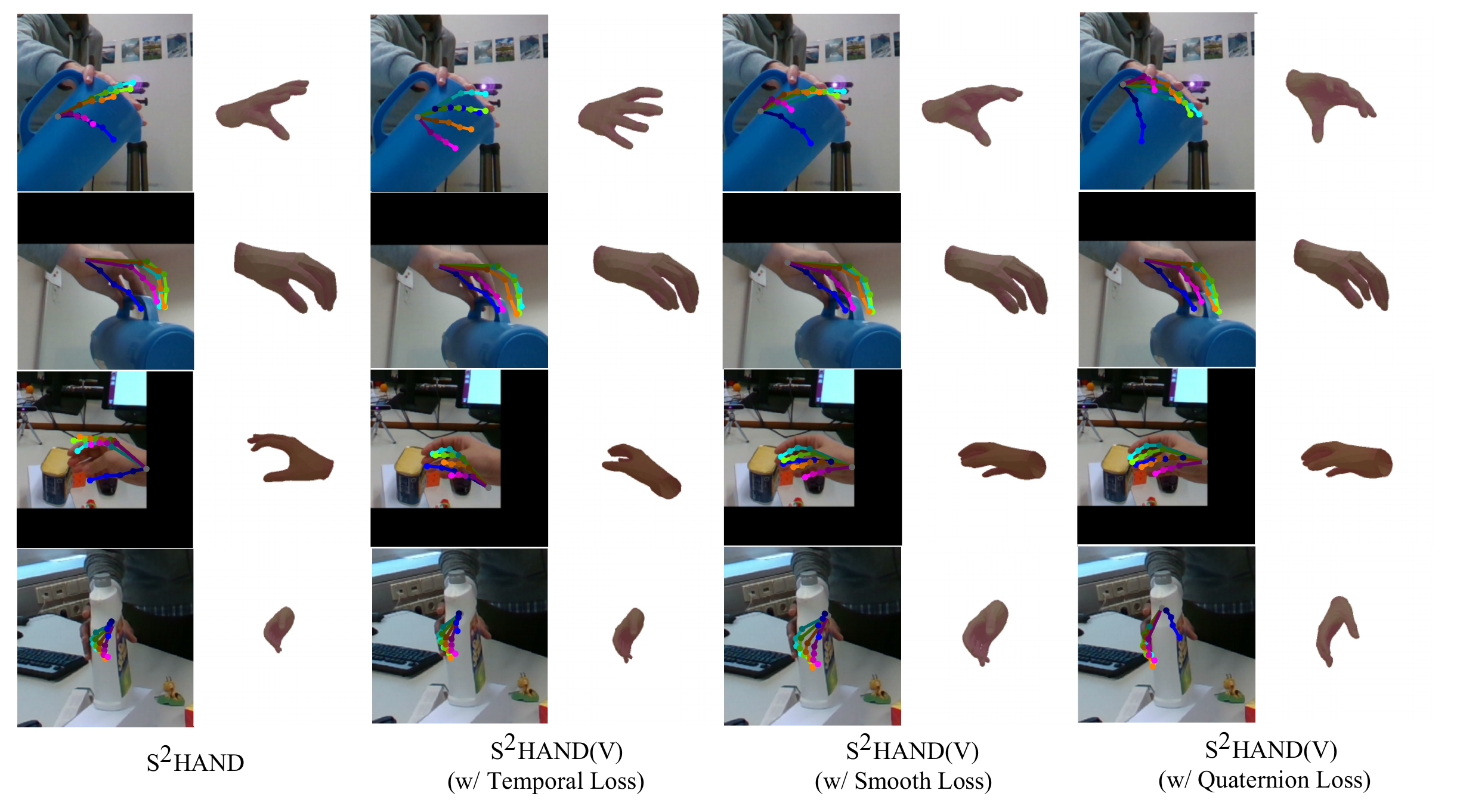}
\caption{Qualitative comparison of different motion-related constraints on the HO-3D testing set. Our ${\rm {S}^{2}HAND(V)}$ with the quaternion loss achieves the best qualitative results.
}
\label{fig:comparision}
%\label{fig:onecol}
\vspace{-0.5cm}
\end{figure*}
Then, we make comparison of different 2D keypoint sets. In our approach, there are three sets of 2D keypoints, including detected keypoints $J^{de}$, estimated 2D keypoints $J^{2d}$, and output projected keypoints $J^{pro}$, where $J^{de}$ is used as supervision terms while $J^{2d}$ and $J^{pro}$ are output items.
In our setting, we use multiple 2D representations to boost the final 3D hand reconstruction, so we do not advocate the novelty of 2D hand estimation, but compare 2D accuracy in the training set to demonstrate the effect of learning from noisy supervision and the benefits of the proposed 2D-3D consistency. Although we use OpenPose outputs as the keypoint supervision source (see \textbf{\textit{OpenPose}} in Fig.~\ref{fig:2d}), we get lower overall 2D MPJPE when we pre-train the 2D and 3D branches separately (see \textbf{\textit{Predicted w/o 2D-3D}} and \textbf{\textit{Projected w/o 2D-3D}} in Fig.~\ref{fig:2d}).
After finetuning these two branches with 2D-3D consistency, we find both of them gain additional benefits.
After the finetuning, the 2D branch (\textbf{\textit{Predicted w/ 2D-3D}}) gains 5.4\% reduction in 2D MPJPE and the 3D branch (\textbf{\textit{Projected w/ 2D-3D}}) gains 9.3\% reduction in 2D MPJPE.
From the curves, we can see that 2D keypoint estimation (including OpenPose and our 2D branch) gets higher accuracy in small distance thresholds while the regression-based methods (\textbf{\textit{Projected w/o 2D-3D}}) get higher accuracy with larger distance threshold.
%This is highly consistent with the motivation of this article.
From the curves, the proposed 2D-3D consistency can improve the 3D branch in all distance thresholds, which verifies the rationality of our network design.

Last, we compare the weak-supervised (WSL) scheme using ground truth annotations with our self-supervised (SSL) approach to investigate the ability of our method to handle noisy supervision sources.
Both settings use the same network structure and implementation, and WSL uses the ground truth 2D keypoint annotations whose keypoint confidences are set to be the same.

As shown in Table~\ref{table:ssl-wsl}, our SSL approach has better performance than WSL settings on both datasets.
We think this is because the detection confidence information is embedded into the proposed loss functions, which helps the network discriminate different accuracy in the noisy samples.
In addition, we find that the SSL method outperforms the WSL method in a smaller amplitude on HO-3D (by 1.0\%) than that on FreiHAND (by 4.9\%). We think this is because the HO-3D contains more occluded hands, resulting in poor 2D detection results.
Therefore, we conclude that noisy 2D keypoints can supervise shape learning for the hand reconstruction task, while the quality of the unlabeled image also has a certain impact.

\subsubsection{Consistent Hand Reconstruction from Video Sequences}
\label{HR from sq}
For consistent self-supervised hand reconstruction from video sequences with ${\rm {S}^{2}HAND(V)}$, we conduct ablation studies on HO-3D since it is a widely used challenging dataset that presents hand-object interaction images with sequence information.

We first study the accuracy performance of quaternion loss in comparison with other commonly used motion-related constraints modeling smoothness in sequence outputs. The compared constraints include temporal loss from \cite{liu2021semi} closing neighboring poses as much as possible and smooth loss from \cite{yang2020seqhand} limiting neighboring pose variation with a threshold. Quantitative and qualitative results are presented in Table~\ref{table:Quat comparison} and Fig.~\ref{fig:comparision}.
Compared to ${\rm {S}^{2}HAND}$, ${\rm {S}^{2}HAND(V)}$ with quaternion loss improves single frame prediction by reducing 3.5\% in MPVPE, while ${\rm {S}^{2}HAND(V)}$ with smooth loss from \cite{liu2021semi} only gets a reduction of less than 1\% in MPVPE and ${\rm {S}^{2}HAND(V)}$ with temporal loss from \cite{yang2020seqhand} even degenerates the accuracy.
We think this is because weak supervision is prone to optimize models in the wrong direction and suffers from being too sketchy and limited under self-supervised settings. 
The temporal loss from \cite{yang2020seqhand} is beneficial when the 3D annotation is available but may collapse the model by making the network insensitive to the high-frequency details in absence of strong supervision signals.
The smooth loss from \cite{liu2021semi} introduces threshold, but in exchange enlarges the solution space and vanishes when the threshold is exceeded.
In contrast, quaternion loss narrows the solution space based on hand structures and hand motion dynamics and provides more significant supervision signals, which proves its effectiveness over other similar constraints in accuracy.
Note that occlusion can be properly resolved during inference in Fig.~\ref{fig:comparision}, we believe it learned from the detection confidence in Eq.~\ref{eq:geo_s} and the constraints from motion in Eq.~\ref{eq:quat}.   

\begin{table}[t]
\vspace{0.2cm}
\begin{minipage}{\linewidth}
\centering
\small
{\def\arraystretch{1} \tabcolsep=0.75em 
\caption{Comparison of different configurations of the quaternion loss on the HO-3D testing set.}
\label{table:Quat self comparison}
\begin{tabular}{c|cc|cc}
        \hline
        %Supervisions & Pose \quad\quad & Mesh \quad & F-score & F-score\\
        %Terms & Error \downarrow & Error \downarrow & @5mm \uparrow & @15mm \uparrow\\
        Config & $\rm {AUC}_{J}$$\uparrow$ & MPJPE$\downarrow$& $\rm {AUC}_{V}$$\uparrow$ & MPVPE$\downarrow$ \\%& STA MPJPE* $\downarrow$
        \hline
        interv=1, n=3 & 0.778 & 1.11 & 0.780 & 1.10 \\%2 (3D branch) & 4.81 \textcolor{red}{to-update}
        % & Ours & 4.73 & 0.772 & 1.14 & 0.775 & 1.13 & 0.45 & 0.93 \\%7
        % & Ours (w/ ref.)\\
        interv=3, n=3 & \textbf{0.780}  & \textbf{1.10} & \textbf{0.782} & \textbf{1.09}\\
        interv=6, n=3 & 0.775 & 1.13 & 0.776 & 1.12\\
        interv=3, n=6 & 0.774 & 1.13 & 0.777 & 1.12 \\
        \hline
\end{tabular}
}
\end{minipage}
\vspace{-0.7cm}
\end{table}

\begin{figure*}[t]
\vspace{-0.3cm}
\centering
\includegraphics[width=0.99\linewidth]{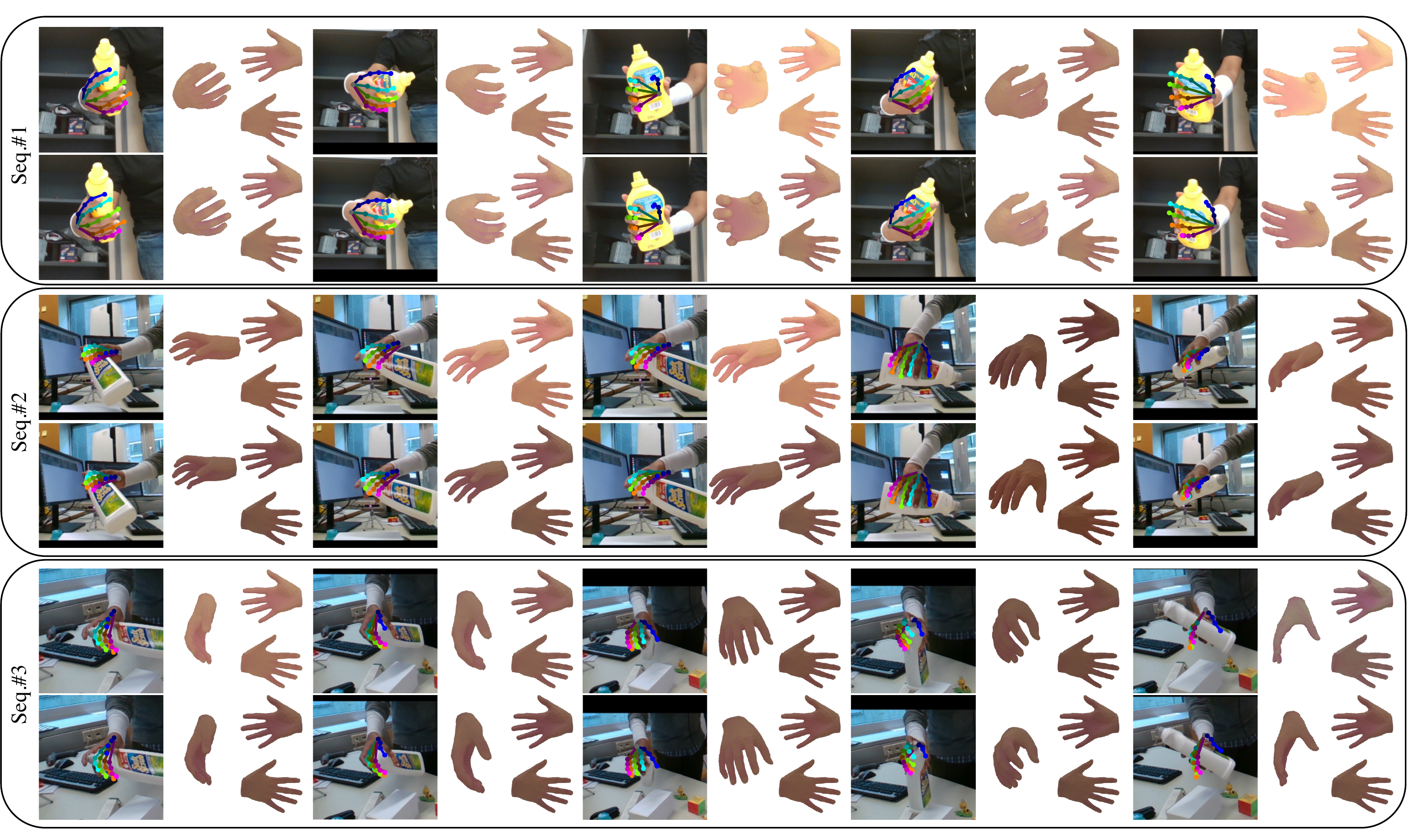}
\caption{
Qualitative demonstration of the effectiveness of the $\rm T\&S$ consistency loss. 
For each frame, we show output keypoints (left), output 3D reconstruction (middle), and both sides of the output textures (with lighting) in flat hands (top-/bottom-right). 
For each sequence, we show results without $\rm T\&S$ loss on the top row and with $\rm T\&S$ loss on the bottom row. 
$\rm T\&S$ loss significantly improves the output appearance consistency in sequence predictions.
}
\label{fig:TS}
%\label{fig:onecol}
\vspace{-0.1cm}
\end{figure*}

\begin{figure*}[t]
\vspace{-0.1cm}
\centering
\includegraphics[width=1\linewidth]{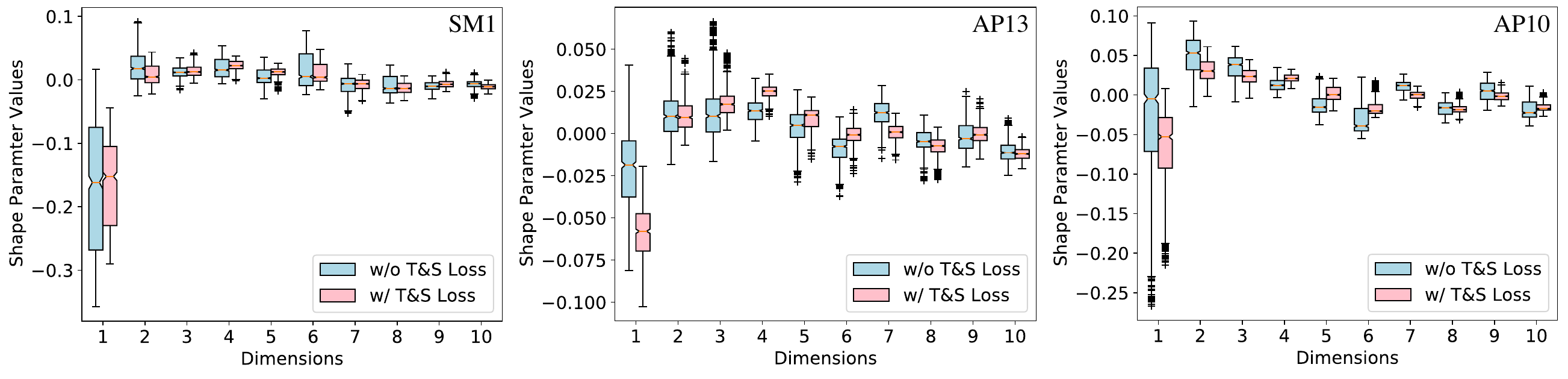}
\caption{Boxplots of shape parameters in sequence predictions on the HO-3D testing set. SM1, AP13, and AP10 are sequences from HO-3D testing set. $\rm T\&S$ loss reduces S.D. across all 10 dimensions of the shape parameters.}
\label{fig:boxplot}
\vspace{-0.6cm}
\end{figure*}

We next explore our quaternion loss with different configurations in terms of the actual frame interval ($\rm interv$) of sampled frames and the number of frames (See $\rm n$ in Eq.~\ref{eq:quat}).
The results are shown in Table~\ref{table:Quat self comparison}.
A medium interval value ($\rm interv = 3$ in Table~\ref{table:Quat self comparison}) achieves the best performance.
We attribute this to two reasons. On one hand, large interval may witness changes of the joint rotation axis in sampled frames, which contradicts the fixed axis prior in quaternion interpolation (See Section~\ref{Sql}). On the other hand, small interval only witnesses small rotation angles, which limits the effect of quaternion loss.
Also, interval is related to the motion speed. In fact, a slow hand motion with larger interval would be equivalent to a fast hand motion with smaller interval. Though the quaternion loss is designed to handle hand motion speed variation, an optimum interval still exists based on the overall motion speed.
For number of frames $\rm n$, increasing it causes model accuracy to drop. We believe this is because optimizing multiple unconfident predictions at the same time puts an extra burden on the gradient-descent-based optimizer and destabilizes the learning procedure. From above, we conclude that the best configuration of quaternion loss depends on the frame rates of input video sequences and the confidence of the output. Adjusting the configuration of quaternion loss dynamically according to the input and the output can be a promising direction for future work.
\begin{figure*}[t]
\vspace{-0.3cm}
\centering
\includegraphics[width=\linewidth]{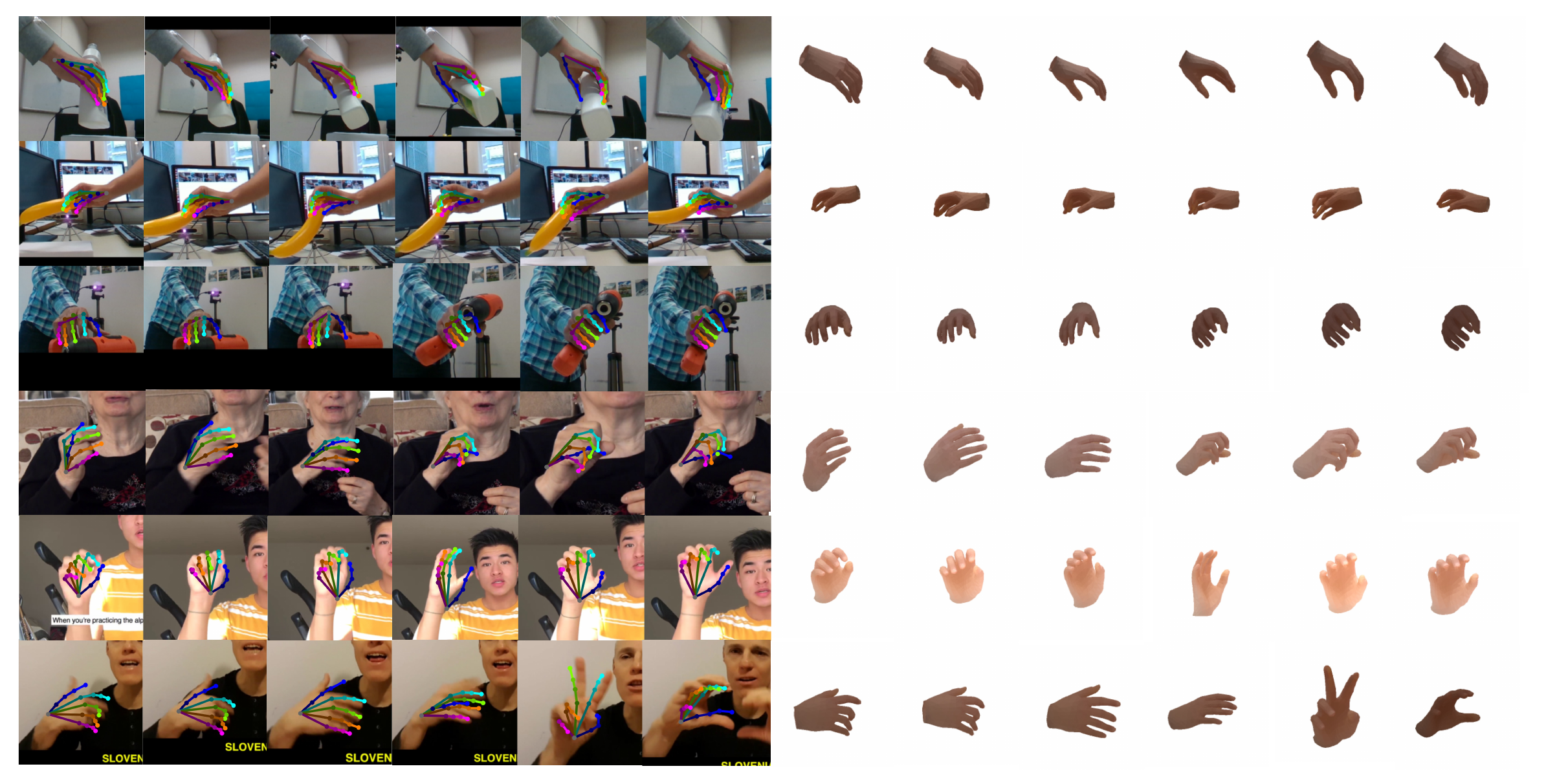}
\caption{Qualitative results of ${\rm {S}^{2}HAND(V)}$ with extra in-the-wild data. The first three rows show results from HO-3D and the last three rows show results from YT 3D.
}
\label{fig:HOYT}
%\label{fig:onecol}
\vspace{-0.05cm}
\end{figure*}

\begin{table*}[ht]
\begin{minipage}{\linewidth}
\centering
\small
{\def\arraystretch{1} \tabcolsep=1.2em 
\caption{Results of absorbing extra in-the-wild data from YT 3D on the HO-3D testing set. * represents changing camera model from perspective to the orthogonal model, which enables to learn with in-the-wild data without camera information.}
\label{table:in the wild}
\begin{tabular}{c|cc|cccc}
        \hline
        %Supervisions & Pose \quad\quad & Mesh \quad & F-score & F-score\\
        %Terms & Error \downarrow & Error \downarrow & @5mm \uparrow & @15mm \uparrow\\
        Method & $\rm {AUC}_{J}$$\uparrow$ & MPJPE$\downarrow$& $\rm {AUC}_{V}$$\uparrow$ & MPVPE$\downarrow$ & ${\rm F}_5$$\uparrow$ & ${\rm F}_{15}$$\uparrow$\\%& STA MPJPE* $\downarrow$
        \hline
        ${\rm {S}^{2}HAND}$ & 0.773 & 1.14 & 0.777 & 1.12 & 0.45 & 0.93\\
        ${\rm {S}^{2}HAND}$* & 0.770  & 1.15 & 0.774 & 1.13 & 0.45 & 0.93\\%2 (3D branch) & 4.81 \textcolor{red}{to-update}
        % & Ours & 4.73 & 0.772 & 1.14 & 0.775 & 1.13 & 0.45 & 0.93 \\%7
        % & Ours (w/ ref.)\\
        ${\rm {S}^{2}HAND(V)}$* (w/ Quaternion Loss) & 0.778 & 1.11 & 0.780  & 1.10 & 0.45 & \textbf{0.94} \\
        %- & Ours (w/ RF) & - & - & - & - & - & - &\\
        ${\rm {S}^{2}HAND(V)}$* (w/ Quaternion Loss) + YT 3D data & \textbf{0.782} & \textbf{1.09} & \textbf{0.783} & \textbf{1.09} & \textbf{0.46} & \textbf{0.94} \\
        %- & Ours (w/ RF) & - & - & - & - & - & - &\\
        \hline
\end{tabular}
}
\end{minipage}
\vspace{-0.1cm}
\end{table*}

We then compare smoothness performance of quaternion loss with others by concatenating their single frame predictions to corresponding sequences predictions.
%The results are
The results are reported on both train set and test set in Table~\ref{table:smooth comparison}.
Note that we only report ACC-ERR on train set since 3D ground truth is required to calculate ACC-ERR.
As shown in Table ~\ref{table:smooth comparison}, all smoothness related prior improve the smoothness of the sequence outputs, and our proposed quaternion loss achieves the best performance in all evaluation matrices.
%\sout{And we observed that method}
\cite{yang2020seqhand} comes second in terms of smoothness, but shows the worst performance on accuracy in Table ~\ref{table:Quat comparison}. We think better smoothness does not imply higher accuracy, where a trade-off between smoothness and accuracy is shown in some cases. %and actually a certain balance exists in some cases.
Our quaternion loss differently does best on both accuracy and smoothness, which again proves its superiority over similar methods.
In addition, we report the average quaternion loss of the concatenated sequences predictions.
We find that our quaternion loss is highly in line with ACC and ACC-ERR, which is encouraging since they are calculated from completely different angles.
ACC-ERR and ACC regard no hand structure and hand motion characteristic, rely solely on mechanics in terms of acceleration, while quaternion loss does the opposite.
Thus, we think our proposed quaternion loss not only proves its advantages over other loss functions in smoothness, but is capable of being the metric measuring smoothness of sequence predictions as well.

\begin{table}[t]
% \begin{minipage}{\linewidth}
\centering
{\def\arraystretch{1} 
\caption{Results of hand appearance consistency of our methods on the HO-3D testing set. Texture S.D. is computed using lighted textures defined in Eq.~\ref{eq:lighted_texture}.}
\label{table:consistency}
\scalebox{0.9}{
\begin{tabular}{c|cc}
        \hline
        Method & Texture S.D. & Shape S.D.\\
        \hline
        ${\rm {S}^{2}HAND}$ & 0.033 & 0.013\\
        ${\rm {S}^{2}HAND(V)}$ (w/ Quaternion Loss) & 0.042 & 0.016\\
        ${\rm {S}^{2}HAND(V)}$ (w/ Quaternion Loss, T\&S Loss) & \textbf{0.016}  & \textbf{0.012}\\
        \hline
\end{tabular}
}
}
% \end{minipage}
\vspace{-0.2cm}
\end{table}

Finally, we inspect the effect of regularizing outputs of hand shape and texture in sequence predictions with the T\&S consistency loss.
The results are presented in Table~\ref{table:Quat comparison}, Table~\ref{table:consistency}, Fig.~\ref{fig:TS} and Fig.~\ref{fig:boxplot}. Though the proposed $\rm T\&S$ consistency loss does not further improve pose accuracy as shown in Table~\ref{table:Quat comparison}, it significantly improves the hand appearance consistency in sequence predictions. Quantitatively, with the $\rm T\&S$ loss, the shape S.D. and texture S.D. drops 62\% and 25\% respectively (see Table~\ref{table:consistency}). Qualitatively, ${\rm {S}^{2}HAND(V)}$ with the $\rm T\&S$ loss gives more consistent reconstructions of different frames from the same sequence than without $\rm T\&S$ as shown in Fig.~\ref{fig:TS}. We also provide boxplots of shape parameters to show dimensional S.D. reduction in Fig.~\ref{fig:boxplot}. Additionally, a visualization of per-face texture S.D. is provided in the supplementary material.

\subsubsection{Hand Reconstruction with Extra In-the-wild Data}
\label{HR in the wild}
For hand reconstruction with extra in-the-wild data to fully exploit the advantage of our self-supervised method ${\rm {S}^{2}HAND(V)}$, we use 34 train videos from YT 3D \cite{kulon2020weakly} and switch to orthogonal camera model with corresponding camera projection to enable learning with in-the-wild data without camera information.

The results are presented in Table~\ref{table:in the wild} and Fig.~\ref{fig:HOYT}.
Changing the camera model makes the model performance drop a little, which may be caused by the ambiguity of camera focal length.
Then, imposing proposed quaternion loss boosts the performance by 3.4\%, which conforms to experiments results in Section~\ref{HR from sq}. Finally, adding extra in-the-wild data further improves our model by 1.8\%, resulting in 1.09cm in MPVPE, which is the best result we can get on the HO-3D test set. From above, we see that our proposed method is able to utilize extra in-the-wild data without any camera information and benefits from these training data. It is worth noticing that there is a certain domain gap between HO-3D and YT 3D since HO-3D regards hand-object interaction scenario while YT 3D is mostly comprised of sign language videos. However, the improvement on HO-3D has still been witnessed, which we think proves the significance of utilizing in the data and the advantage of the proposed method.

\section{Conclusion}
In this work, we have proposed self-supervised 3D hand reconstruction models ${\rm {S}^{2}HAND}$ and ${\rm {S}^{2}HAND(V)}$ which can be trained from a collection of hand images and video sequences without manual annotations, respectively. The 3D hand reconstruction network in both models encodes the input image into a set of meaningful semantic parameters that represent hand pose, shape, texture, illumination, and the camera viewpoint. These parameters can be decoded into a textured 3D hand mesh as well as a set of 3D joints, and in turn, the 3D mesh and joints can be projected into the 2D image space, which enables our network to be end-to-end learned.
We further exploit the self-supervision signals embedded in hand motion videos by developing a novel quaternion loss and a texture and shape consistency loss to obtain more accurate and consistent hand reconstruction.
Experimental results show that our models perform well under noisy supervision sources captured from 2D hand keypoint detection, and achieve comparable performance to the state-of-the-art fully-supervised method.
Moreover, the experiments on in-the-wild video data show that our self-supervised model is effective to learn useful information from in-the-wild data to further improve its performance. 

For the future study, we think the texture and shape representation could be investigated deeply. Notice that we only estimate a per-face color and adopt a low-resolution hand model with 778 vertices. Besides, we rely on the shape and texture regularization terms to learn hand appearance from limited raw data under complex environments and potential hand-object interaction. These defects lead to the reconstruction results having limited details. Recently, Corona et al. \cite{corona2022lisa} utilize implicit representation to learn hand shape and texture of high quality. This points in a good direction for further exploration.

\ifCLASSOPTIONcompsoc

\section*{Acknowledgment}
This work was supported by the Natural Science Fund for Distinguished Young Scholars of Hubei Province under Grant 2022CFA075, the National Natural Science Foundation of China under Grant 62106177, and the National Science Fund for Distinguished Young Scholars of China under Grant 41725005. The numerical calculation was supported by the supercomputing system in the Super-computing Center of Wuhan University.

\ifCLASSOPTIONcaptionsoff
  \newpage
\fi

\bibliographystyle{IEEEtran}

\bibliography{hzsbib.bib}

\begin{thebibliography}{10}
\providecommand{\url}[1]{#1}
\csname url@samestyle\endcsname
\providecommand{\newblock}{\relax}
\providecommand{\bibinfo}[2]{#2}
\providecommand{\BIBentrySTDinterwordspacing}{\spaceskip=0pt\relax}
\providecommand{\BIBentryALTinterwordstretchfactor}{4}
\providecommand{\BIBentryALTinterwordspacing}{\spaceskip=\fontdimen2\font plus
\BIBentryALTinterwordstretchfactor\fontdimen3\font minus
  \fontdimen4\font\relax}
\providecommand{\BIBforeignlanguage}[2]{{%
\expandafter\ifx\csname l@#1\endcsname\relax
\typeout{** WARNING: IEEEtran.bst: No hyphenation pattern has been}%
\typeout{** loaded for the language `#1'. Using the pattern for}%
\typeout{** the default language instead.}%
\else
\language=\csname l@#1\endcsname
\fi
#2}}
\providecommand{\BIBdecl}{\relax}
\BIBdecl

\bibitem{lee2011two}
H.~Lee, M.~Billinghurst, and W.~Woo, ``Two-handed tangible interaction
  techniques for composing augmented blocks,'' \emph{Virtual Reality}, vol.~15,
  no.~2, pp. 133--146, 2011.

\bibitem{camgoz2018neural}
N.~C. Camgoz, S.~Hadfield, O.~Koller, H.~Ney, and R.~Bowden, ``Neural sign
  language translation,'' in \emph{Conference on Computer Vision and Pattern
  Recognition}, 2018.

\bibitem{camgoz2020sign}
N.~C. Camgoz, O.~Koller, S.~Hadfield, and R.~Bowden, ``Sign language
  transformers: Joint end-to-end sign language recognition and translation,''
  in \emph{Conference on Computer Vision and Pattern Recognition}, 2020.

\bibitem{holl2018efficient}
M.~H{\"o}ll, M.~Oberweger, C.~Arth, and V.~Lepetit, ``Efficient physics-based
  implementation for realistic hand-object interaction in virtual reality,'' in
  \emph{Conference on Virtual Reality and 3D User Interfaces}, 2018.

\bibitem{parelli2020exploiting}
M.~Parelli, K.~Papadimitriou, G.~Potamianos, G.~Pavlakos, and P.~Maragos,
  ``Exploiting 3d hand pose estimation in deep learning-based sign language
  recognition from rgb videos,'' in \emph{European Conference on Computer
  Vision}, 2020.

\bibitem{tu2019action}
Z.~Tu, H.~Li, D.~Zhang, J.~Dauwels, B.~Li, and J.~Yuan, ``Action-stage
  emphasized spatiotemporal vlad for video action recognition,'' \emph{IEEE
  Transactions on Image Processing}, vol.~28, no.~6, pp. 2799--2812, 2019.

\bibitem{ge2018robust}
L.~Ge, H.~Liang, J.~Yuan, and D.~Thalmann, ``Robust 3d hand pose estimation
  from single depth images using multi-view cnns,'' \emph{IEEE Transactions on
  Image Processing}, vol.~27, no.~9, pp. 4422--4436, 2018.

\bibitem{yu2020humbi}
Z.~Yu, J.~S. Yoon, I.~K. Lee, P.~Venkatesh, J.~Park, J.~Yu, and H.~S. Park,
  ``Humbi: A large multiview dataset of human body expressions,'' in
  \emph{Conference on Computer Vision and Pattern Recognition}, 2020.

\bibitem{zhao2020hand}
Z.~Zhao, T.~Wang, S.~Xia, and Y.~Wang, ``Hand-3d-studio: A new multi-view
  system for 3d hand reconstruction,'' in \emph{IEEE International Conference
  on Acoustics, Speech, and Signal Processing}, 2020.

\bibitem{poier2018learning}
G.~Poier, D.~Schinagl, and H.~Bischof, ``Learning pose specific representations
  by predicting different views,'' in \emph{Conference on Computer Vision and
  Pattern Recognition}, 2018.

\bibitem{armagan2020measuring}
A.~Armagan, G.~Garcia-Hernando, S.~Baek, S.~Hampali, M.~Rad, Z.~Zhang, S.~Xie,
  M.~Chen, B.~Zhang, F.~Xiong \emph{et~al.}, ``Measuring generalisation to
  unseen viewpoints, articulations, shapes and objects for 3d hand pose
  estimation under hand-object interaction,'' in \emph{European Conference on
  Computer Vision}, 2020.

\bibitem{chen2019so}
Y.~Chen, Z.~Tu, L.~Ge, D.~Zhang, R.~Chen, and J.~Yuan, ``So-handnet:
  Self-organizing network for 3d hand pose estimation with semi-supervised
  learning,'' in \emph{International Conference on Computer Vision}, 2019.

\bibitem{ge2016robust}
L.~Ge, H.~Liang, J.~Yuan, and D.~Thalmann, ``Robust 3d hand pose estimation in
  single depth images: from single-view cnn to multi-view cnns,'' in
  \emph{Conference on Computer Vision and Pattern Recognition}, 2016.

\bibitem{huang2020hand}
L.~Huang, J.~Tan, J.~Liu, and J.~Yuan, ``Hand-transformer: Non-autoregressive
  structured modeling for 3d hand pose estimation,'' in \emph{European
  Conference on Computer Vision}, 2020.

\bibitem{yuan2018depth}
S.~Yuan, G.~Garcia-Hernando, B.~Stenger, G.~Moon, J.~Yong~Chang, K.~Mu~Lee,
  P.~Molchanov, J.~Kautz, S.~Honari, L.~Ge \emph{et~al.}, ``Depth-based 3d hand
  pose estimation: From current achievements to future goals,'' in
  \emph{Conference on Computer Vision and Pattern Recognition}, 2018.

\bibitem{athitsos2003estimating}
V.~Athitsos and S.~Sclaroff, ``Estimating 3d hand pose from a cluttered
  image,'' in \emph{Conference on Computer Vision and Pattern Recognition},
  2003.

\bibitem{cai2018weakly}
Y.~Cai, L.~Ge, J.~Cai, and J.~Yuan, ``Weakly-supervised 3d hand pose estimation
  from monocular rgb images,'' in \emph{European Conference on Computer
  Vision}, 2018.

\bibitem{iqbal2018hand}
U.~Iqbal, P.~Molchanov, T.~Breuel Juergen~Gall, and J.~Kautz, ``Hand pose
  estimation via latent 2.5 d heatmap regression,'' in \emph{European
  Conference on Computer Vision}, 2018.

\bibitem{spurr2020weakly}
A.~Spurr, U.~Iqbal, P.~Molchanov, O.~Hilliges, and J.~Kautz, ``Weakly
  supervised 3d hand pose estimation via biomechanical constraints,'' in
  \emph{European Conference on Computer Vision}, 2020.

\bibitem{zimmermann2017learning}
C.~Zimmermann and T.~Brox, ``Learning to estimate 3d hand pose from single rgb
  images,'' in \emph{International Conference on Computer Vision}, 2017.

\bibitem{Ge_2019_CVPR}
L.~Ge, Z.~Ren, Y.~Li, Z.~Xue, Y.~Wang, J.~Cai, and J.~Yuan, ``3d hand shape and
  pose estimation from a single rgb image,'' in \emph{Conference on Computer
  Vision and Pattern Recognition}, 2019.

\bibitem{Kulon_2020_CVPR}
D.~Kulon, R.~A. Guler, I.~Kokkinos, M.~M. Bronstein, and S.~Zafeiriou,
  ``Weakly-supervised mesh-convolutional hand reconstruction in the wild,'' in
  \emph{Conference on Computer Vision and Pattern Recognition}, 2020.

\bibitem{hasson2020leveraging}
Y.~Hasson, B.~Tekin, F.~Bogo, I.~Laptev, M.~Pollefeys, and C.~Schmid,
  ``Leveraging photometric consistency over time for sparsely supervised
  hand-object reconstruction,'' in \emph{Conference on Computer Vision and
  Pattern Recognition}, 2020.

\bibitem{hasson2019learning}
Y.~Hasson, G.~Varol, D.~Tzionas, I.~Kalevatykh, M.~J. Black, I.~Laptev, and
  C.~Schmid, ``Learning joint reconstruction of hands and manipulated
  objects,'' in \emph{Conference on Computer Vision and Pattern Recognition},
  2019.

\bibitem{qian2020parametric}
N.~Qian, J.~Wang, F.~Mueller, F.~Bernard, V.~Golyanik, and C.~Theobalt,
  ``Parametric hand texture model for 3d hand reconstruction and
  personalization,'' in \emph{European Conference on Computer Vision}, 2020.

\bibitem{cao2019openpose}
Z.~Cao, T.~Simon, S.-E. Wei, and Y.~Sheikh, ``Openpose: Realtime multi-person
  2d pose estimation using part affinity fields.'' \emph{IEEE Transactions on
  Pattern Analysis and Machine Intelligence}, vol.~43, no.~01, pp. 172--186,
  2019.

\bibitem{chen2020self}
Y.~Chen, F.~Wu, Z.~Wang, Y.~Song, Y.~Ling, and L.~Bao, ``Self-supervised
  learning of detailed 3d face reconstruction,'' \emph{IEEE Transactions on
  Image Processing}, vol.~29, pp. 8696--8705, 2020.

\bibitem{tewari2017mofa}
A.~Tewari, M.~Zollhofer, H.~Kim, P.~Garrido, F.~Bernard, P.~Perez, and
  C.~Theobalt, ``Mofa: Model-based deep convolutional face autoencoder for
  unsupervised monocular reconstruction,'' in \emph{International Conference on
  Computer Vision Workshops}, 2017.

\bibitem{chen2021model}
Y.~Chen, Z.~Tu, D.~Kang, L.~Bao, Y.~Zhang, X.~Zhe, R.~Chen, and J.~Yuan,
  ``Model-based 3d hand reconstruction via self-supervised learning,'' in
  \emph{Conference on Computer Vision and Pattern Recognition}, 2021.

\bibitem{cai20203d}
Y.~Cai, L.~Ge, J.~Cai, N.~M. Thalmann, and J.~Yuan, ``3d hand pose estimation
  using synthetic data and weakly labeled rgb images,'' \emph{IEEE transactions
  on pattern analysis and machine intelligence}, vol.~43, no.~11, pp.
  3739--3753, 2020.

\bibitem{yang2019disentangling}
L.~Yang and A.~Yao, ``Disentangling latent hands for image synthesis and pose
  estimation,'' in \emph{Conference on Computer Vision and Pattern
  Recognition}, 2019.

\bibitem{lin2021end}
K.~Lin, L.~Wang, and Z.~Liu, ``End-to-end human pose and mesh reconstruction
  with transformers,'' in \emph{Conference on Computer Vision and Pattern
  Recognition}, 2021.

\bibitem{chen2021i2uv}
P.~Chen, Y.~Chen, D.~Yang, F.~Wu, Q.~Li, Q.~Xia, and Y.~Tan, ``I2uv-handnet:
  Image-to-uv prediction network for accurate and high-fidelity 3d hand mesh
  modeling,'' in \emph{International Conference on Computer Vision}, 2021.

\bibitem{liu2021semi}
S.~Liu, H.~Jiang, J.~Xu, S.~Liu, and X.~Wang, ``Semi-supervised 3d hand-object
  poses estimation with interactions in time,'' in \emph{Conference on Computer
  Vision and Pattern Recognition}, 2021.

\bibitem{zhou2021monocular}
Y.~Zhou, M.~Habermann, I.~Habibie, A.~Tewari, C.~Theobalt, and F.~Xu,
  ``Monocular real-time full body capture with inter-part correlations,'' in
  \emph{Conference on Computer Vision and Pattern Recognition}, 2021.

\bibitem{chen2021camera}
X.~Chen, Y.~Liu, C.~Ma, J.~Chang, H.~Wang, T.~Chen, X.~Guo, P.~Wan, and
  W.~Zheng, ``Camera-space hand mesh recovery via semantic aggregation and
  adaptive 2d-1d registration,'' in \emph{Conference on Computer Vision and
  Pattern Recognition}, 2021.

\bibitem{cao2021reconstructing}
Z.~Cao, I.~Radosavovic, A.~Kanazawa, and J.~Malik, ``Reconstructing hand-object
  interactions in the wild,'' in \emph{International Conference on Computer
  Vision}, 2021.

\bibitem{kulon2020weakly}
D.~Kulon, R.~A. Guler, I.~Kokkinos, M.~M. Bronstein, and S.~Zafeiriou,
  ``Weakly-supervised mesh-convolutional hand reconstruction in the wild,'' in
  \emph{Conference on Computer Vision and Pattern Recognition}, 2020.

\bibitem{lim2018simple}
I.~Lim, A.~Dielen, M.~Campen, and L.~Kobbelt, ``A simple approach to intrinsic
  correspondence learning on unstructured 3d meshes,'' in \emph{European
  Conference on Computer Vision}, 2018.

\bibitem{dkulon2019rec}
D.~Kulon, H.~Wang, R.~A. G{\"{u}}ler, M.~M. Bronstein, and S.~Zafeiriou,
  ``Single image 3d hand reconstruction with mesh convolutions,'' in
  \emph{British Machine Vision Conference}, 2019.

\bibitem{defferrard2016convolutional}
M.~Defferrard, X.~Bresson, and P.~Vandergheynst, ``Convolutional neural
  networks on graphs with fast localized spectral filtering,'' \emph{Advances
  in neural information processing systems}, vol.~29, p. 3844–3852, 2016.

\bibitem{ge20193d}
L.~Ge, Z.~Ren, Y.~Li, Z.~Xue, Y.~Wang, J.~Cai, and J.~Yuan, ``3d hand shape and
  pose estimation from a single rgb image,'' in \emph{Conference on Computer
  Vision and Pattern Recognition}, 2019.

\bibitem{MANO:SIGGRAPHASIA:2017}
J.~Romero, D.~Tzionas, and M.~J. Black, ``Embodied hands: Modeling and
  capturing hands and bodies together,'' \emph{ACM Transactions on Graphics
  (Proceedings of SIGGRAPH Asia)}, vol.~36, no.~6, 2017.

\bibitem{boukhayma20193d}
A.~Boukhayma, R.~d. Bem, and P.~H. Torr, ``3d hand shape and pose from images
  in the wild,'' in \emph{Conference on Computer Vision and Pattern
  Recognition}, 2019.

\bibitem{chen2021joint}
Y.~Chen, Z.~Tu, D.~Kang, R.~Chen, L.~Bao, Z.~Zhang, and J.~Yuan, ``Joint
  hand-object 3d reconstruction from a single image with cross-branch feature
  fusion,'' \emph{IEEE Transactions on Image Processing}, vol.~30, pp.
  4008--4021, 2021.

\bibitem{zimmermann2019freihand}
C.~Zimmermann, D.~Ceylan, J.~Yang, B.~Russell, M.~Argus, and T.~Brox,
  ``Freihand: A dataset for markerless capture of hand pose and shape from
  single rgb images,'' in \emph{International Conference on Computer Vision},
  2019.

\bibitem{Baek_2019_CVPR}
S.~Baek, K.~I. Kim, and T.-K. Kim, ``Pushing the envelope for rgb-based dense
  3d hand pose estimation via neural rendering,'' in \emph{Conference on
  Computer Vision and Pattern Recognition}, 2019.

\bibitem{baek2020weakly}
------, ``Weakly-supervised domain adaptation via gan and mesh model for
  estimating 3d hand poses interacting objects,'' in \emph{Conference on
  Computer Vision and Pattern Recognition}, 2020.

\bibitem{wang_SIGAsia2020}
J.~Wang, F.~Mueller, F.~Bernard, S.~Sorli, O.~Sotnychenko, N.~Qian, M.~A.
  Otaduy, D.~Casas, and C.~Theobalt, ``Rgb2hands: real-time tracking of 3d hand
  interactions from monocular rgb video,'' \emph{ACM Transactions on Graphics},
  vol.~39, no.~6, pp. 1--16, 2020.

\bibitem{zhou2020monocular}
Y.~Zhou, M.~Habermann, W.~Xu, I.~Habibie, C.~Theobalt, and F.~Xu, ``Monocular
  real-time hand shape and motion capture using multi-modal data,'' in
  \emph{Conference on Computer Vision and Pattern Recognition}, 2020.

\bibitem{panteleris2018using}
P.~Panteleris, I.~Oikonomidis, and A.~Argyros, ``Using a single rgb frame for
  real time 3d hand pose estimation in the wild,'' in \emph{Winter Conference
  on Applications of Computer Vision}, 2018.

\bibitem{Wu_2020_CVPR}
S.~Wu, C.~Rupprecht, and A.~Vedaldi, ``Unsupervised learning of probably
  symmetric deformable 3d objects from images in the wild,'' in
  \emph{Conference on Computer Vision and Pattern Recognition}, 2020.

\bibitem{goel2020shape}
S.~Goel, A.~Kanazawa, and J.~Malik, ``Shape and viewpoint without keypoints,''
  in \emph{European Conference on Computer Vision}, 2020.

\bibitem{spurr2021self}
A.~Spurr, A.~Dahiya, X.~Wang, X.~Zhang, and O.~Hilliges, ``Self-supervised 3d
  hand pose estimation from monocular rgb via contrastive learning,'' in
  \emph{International Conference on Computer Vision}, 2021.

\bibitem{guo2020graph}
S.~Guo, E.~Rigall, L.~Qi, X.~Dong, H.~Li, and J.~Dong, ``Graph-based cnns with
  self-supervised module for 3d hand pose estimation from monocular rgb,''
  \emph{IEEE Transactions on Circuits and Systems for Video Technology},
  vol.~31, no.~4, pp. 1514--1525, 2020.

\bibitem{wan2019self}
C.~Wan, T.~Probst, L.~V. Gool, and A.~Yao, ``Self-supervised 3d hand pose
  estimation through training by fitting,'' in \emph{Conference on Computer
  Vision and Pattern Recognition}, 2019.

\bibitem{tewari2018self}
A.~Tewari, M.~Zollh{\"o}fer, P.~Garrido, F.~Bernard, H.~Kim, P.~P{\'e}rez, and
  C.~Theobalt, ``Self-supervised multi-level face model learning for monocular
  reconstruction at over 250 hz,'' in \emph{Conference on Computer Vision and
  Pattern Recognition}, 2018.

\bibitem{blanz1999morphable}
V.~Blanz and T.~Vetter, ``A morphable model for the synthesis of 3d faces,'' in
  \emph{ACM Transactions on Graphics (Proceedings of SIGGRAPH)}, 1999, p.
  187–194.

\bibitem{de2011model}
M.~de~La~Gorce, D.~J. Fleet, and N.~Paragios, ``Model-based 3d hand pose
  estimation from monocular video,'' \emph{IEEE Transactions on Pattern
  Analysis and Machine Intelligence}, vol.~33, no.~9, pp. 1793--1805, 2011.

\bibitem{de2008model}
M.~de~La~Gorce, N.~Paragios, and D.~J. Fleet, ``Model-based hand tracking with
  texture, shading and self-occlusions,'' in \emph{Conference on Computer
  Vision and Pattern Recognition}, 2008.

\bibitem{cai2019exploiting}
Y.~Cai, L.~Ge, J.~Liu, J.~Cai, T.-J. Cham, J.~Yuan, and N.~M. Thalmann,
  ``Exploiting spatial-temporal relationships for 3d pose estimation via graph
  convolutional networks,'' in \emph{International Conference on Computer
  Vision}, 2019.

\bibitem{chen2021temporal}
L.~Chen, S.-Y. Lin, Y.~Xie, Y.-Y. Lin, and X.~Xie, ``Temporal-aware
  self-supervised learning for 3d hand pose and mesh estimation in videos,'' in
  \emph{Winter Conference on Applications of Computer Vision}, 2021.

\bibitem{yang2020seqhand}
J.~Yang, H.~J. Chang, S.~Lee, and N.~Kwak, ``Seqhand: Rgb-sequence-based 3d
  hand pose and shape estimation,'' in \emph{European Conference on Computer
  Vision}, 2020.

\bibitem{fragkiadaki2015recurrent}
K.~Fragkiadaki, S.~Levine, P.~Felsen, and J.~Malik, ``Recurrent network models
  for human dynamics,'' in \emph{International Conference on Computer Vision},
  2015.

\bibitem{martinez2017human}
J.~Martinez, M.~J. Black, and J.~Romero, ``On human motion prediction using
  recurrent neural networks,'' in \emph{Conference on Computer Vision and
  Pattern Recognition}, 2017.

\bibitem{aksan2019structured}
E.~Aksan, M.~Kaufmann, and O.~Hilliges, ``Structured prediction helps 3d human
  motion modelling,'' in \emph{International Conference on Computer Vision},
  2019.

\bibitem{hernandez2019human}
A.~Hernandez, J.~Gall, and F.~Moreno-Noguer, ``Human motion prediction via
  spatio-temporal inpainting,'' in \emph{International Conference on Computer
  Vision}, 2019.

\bibitem{tan2019efficientnet}
M.~Tan and Q.~V. Le, ``Efficientnet: Rethinking model scaling for convolutional
  neural networks,'' in \emph{International Conference on Machine Learning},
  2019.

\bibitem{kato2018neural}
H.~Kato, Y.~Ushiku, and T.~Harada, ``Neural 3d mesh renderer,'' in
  \emph{Conference on Computer Vision and Pattern Recognition}, 2018.

\bibitem{newell2016stacked}
A.~Newell, K.~Yang, and J.~Deng, ``Stacked hourglass networks for human pose
  estimation,'' in \emph{European Conference on Computer Vision}, 2016.

\bibitem{sun2018integral}
X.~Sun, B.~Xiao, F.~Wei, S.~Liang, and Y.~Wei, ``Integral human pose
  regression,'' in \emph{European Conference on Computer Vision}, 2018.

\bibitem{huber1992robust}
P.~J. Huber, ``Robust estimation of a location parameter,'' in
  \emph{Breakthroughs in statistics}, 1992.

\bibitem{wang2004image}
Z.~Wang, A.~C. Bovik, H.~R. Sheikh, and E.~P. Simoncelli, ``Image quality
  assessment: from error visibility to structural similarity,'' \emph{IEEE
  Transactions on Image Processing}, vol.~13, no.~4, pp. 600--612, 2004.

\bibitem{pavllo2020modeling}
D.~Pavllo, C.~Feichtenhofer, M.~Auli, and D.~Grangier, ``Modeling human motion
  with quaternion-based neural networks,'' \emph{International Journal of
  Computer Vision}, vol. 128, no.~4, pp. 855--872, 2020.

\bibitem{zhang2020quaternion}
X.~Zhang, S.~Qin, Y.~Xu, and H.~Xu, ``Quaternion product units for deep
  learning on 3d rotation groups,'' in \emph{Conference on Computer Vision and
  Pattern Recognition}, 2020.

\bibitem{shoemake1985animating}
K.~Shoemake, ``Animating rotation with quaternion curves,'' in \emph{ACM
  Transactions on Graphics (Proceedings of SIGGRAPH)}, vol.~19, no.~3, 1985, p.
  245–254.

\bibitem{zhou2019continuity}
Y.~Zhou, C.~Barnes, J.~Lu, J.~Yang, and H.~Li, ``On the continuity of rotation
  representations in neural networks,'' in \emph{Conference on Computer Vision
  and Pattern Recognition}, 2019.

\bibitem{hampali2020honnotate}
S.~Hampali, M.~Rad, M.~Oberweger, and V.~Lepetit, ``Honnotate: A method for 3d
  annotation of hand and object poses,'' in \emph{Conference on Computer Vision
  and Pattern Recognition}, 2020.

\bibitem{zhang2017hand}
J.~Zhang, J.~Jiao, M.~Chen, L.~Qu, X.~Xu, and Q.~Yang, ``A hand pose tracking
  benchmark from stereo matching,'' in \emph{International Conference on Image
  Processing}, 2017.

\bibitem{knapitsch2017tanks}
A.~Knapitsch, J.~Park, Q.-Y. Zhou, and V.~Koltun, ``Tanks and temples:
  Benchmarking large-scale scene reconstruction,'' \emph{ACM Transactions on
  Graphics}, vol.~36, no.~4, pp. 1--13, 2017.

\bibitem{kanazawa2019learning}
A.~Kanazawa, J.~Y. Zhang, P.~Felsen, and J.~Malik, ``Learning 3d human dynamics
  from video,'' in \emph{Conference on Computer Vision and Pattern
  Recognition}, 2019.

\bibitem{paszke2017automatic}
A.~Paszke, S.~Gross, S.~Chintala, G.~Chanan, E.~Yang, Z.~DeVito, Z.~Lin,
  A.~Desmaison, L.~Antiga, and A.~Lerer, ``Automatic differentiation in
  pytorch,'' in \emph{Neural Information Processing Systems Workshops}, 2017.

\bibitem{kingma2014adam}
D.~P. Kingma and J.~Ba, ``Adam: A method for stochastic optimization,'' in
  \emph{International Conference for Learning Representations}, 2014.

\bibitem{moon2020i2l}
G.~Moon and K.~M. Lee, ``I2l-meshnet: Image-to-lixel prediction network for
  accurate 3d human pose and mesh estimation from a single rgb image,'' in
  \emph{European Conference on Computer Vision}, 2020.

\bibitem{mueller2018ganerated}
F.~Mueller, F.~Bernard, O.~Sotnychenko, D.~Mehta, S.~Sridhar, D.~Casas, and
  C.~Theobalt, ``Ganerated hands for real-time 3d hand tracking from monocular
  rgb,'' in \emph{Conference on Computer Vision and Pattern Recognition}, 2018.

\bibitem{spurr2018cross}
A.~Spurr, J.~Song, S.~Park, and O.~Hilliges, ``Cross-modal deep variational
  hand pose estimation,'' in \emph{Conference on Computer Vision and Pattern
  Recognition}, 2018.

\bibitem{theodoridis2020cross}
T.~Theodoridis, T.~Chatzis, V.~Solachidis, K.~Dimitropoulos, and P.~Daras,
  ``Cross-modal variational alignment of latent spaces,'' in \emph{Conference
  on Computer Vision and Pattern Recognition Workshops}, 2020.

\bibitem{corona2022lisa}
E.~Corona, T.~Hodan, M.~Vo, F.~Moreno-Noguer, C.~Sweeney, R.~Newcombe, and
  L.~Ma, ``Lisa: Learning implicit shape and appearance of hands,'' in
  \emph{Conference on Computer Vision and Pattern Recognition}, 2022.

\end{thebibliography}

\vspace{-1.4cm}
\begin{IEEEbiography}[{\includegraphics[width=0.95\textwidth]{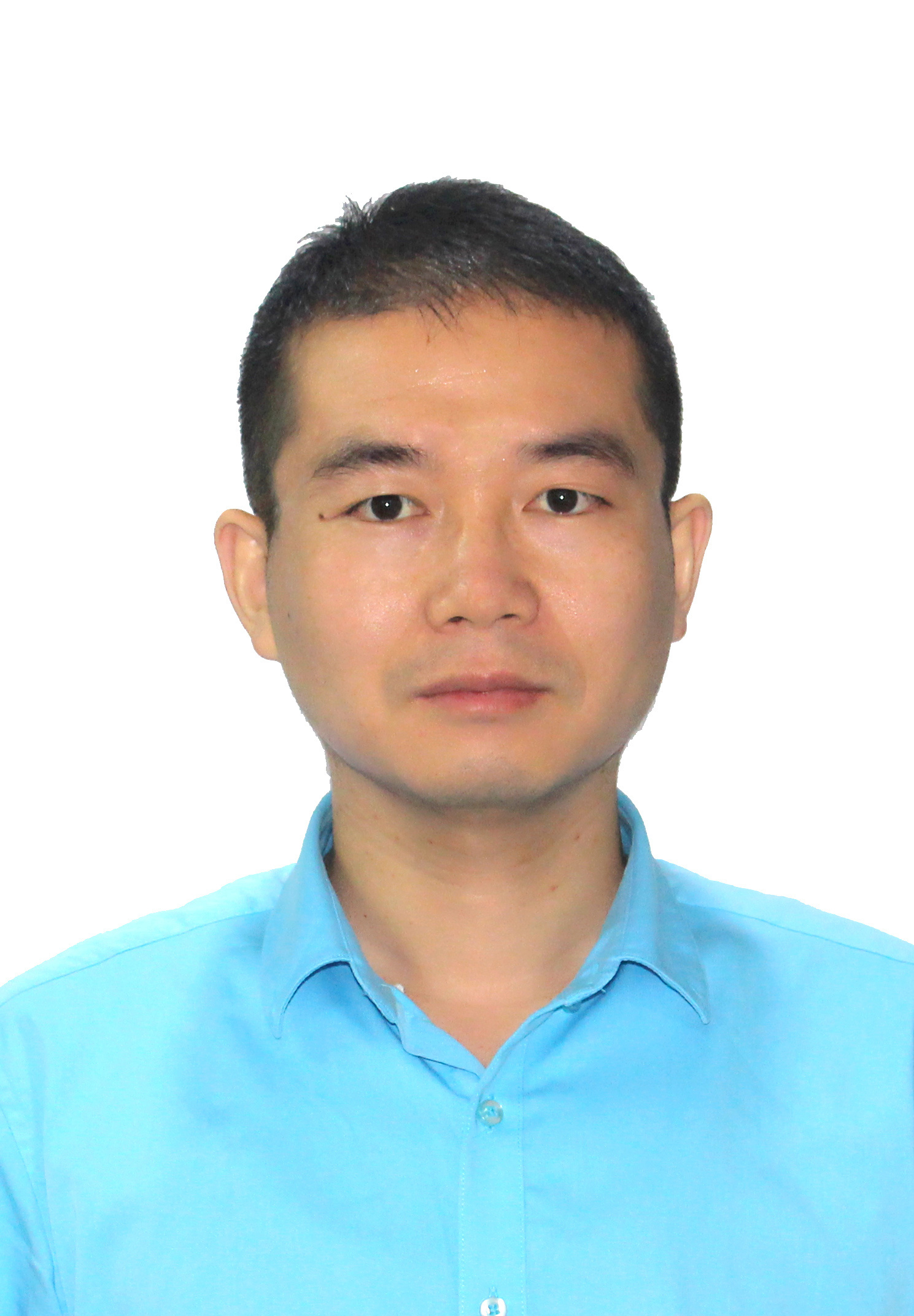}}]
{\textbf{Zhigang Tu}} Received the Ph.D. degree respectively from Wuhan University (China), 2013, and Utrecht University (Netherlands), 2015. From 2015 to 2016, he was a postdoctor at Arizona State University, US. Then from 2016 to 2018, he was a research fellow at Nanyang Technological University, Singapore. He is currently a professor at Wuhan University, China. His research interests include computer vision, machine learning, and video analytics. Special for motion estimation, Human behavior recognition and 3D Reconstruction, and anomaly detection. He has co-/authored more than 60 articles on international SCI-indexed journals and conferences. He is an Associate Editor of TVC (IF=2.835) and a Guest Editor of JVCIR (IF=2.887). He is the first organizer of the ACCV2020 Workshop on MMHAU (Japan). He received the ``Best Student Paper" Award in the $4^{th}$ Asian Conf. Artificial Intelligence Technology.
\end{IEEEbiography}

\vspace{-1.5cm}

\begin{IEEEbiography}[{\includegraphics[width=0.95\textwidth]{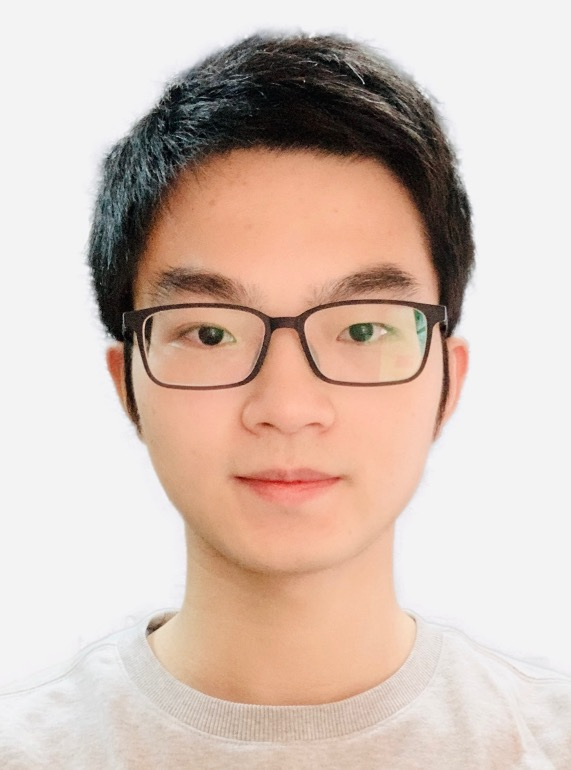}}]
{\textbf{Zhisheng Huang}} received the B.Eng. degree in Surveying and Mapping Engineering Technology from Wuhan University in 2021. He is now a postgraduate student with the State Key Laboratory of Information Engineering in Surveying, Mapping and Remote Sensing, Wuhan University. His research interests mainly include computer vision, machine learning and human-computer interaction.
\end{IEEEbiography}

\vspace{-1.7cm}

\begin{IEEEbiography}[{\includegraphics[width=0.95\textwidth]{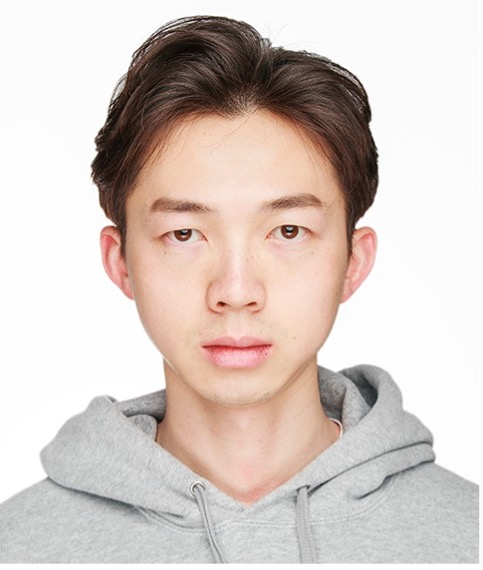}}]
{\textbf{Yujin Chen}} received the B.Eng. and M.Sc. in Geo-Information from Wuhan University. He is currently a Ph.D. student with the Visual Computing Lab, Technical University of Munich, Germany. His research interests include computer vision and machine learning. His current focus is on 3D scene understanding, pose estimation, motion analysis, and representation learning.
\end{IEEEbiography}

\vspace{-1.8cm}

\begin{IEEEbiography}[{\includegraphics[width=0.95\textwidth]{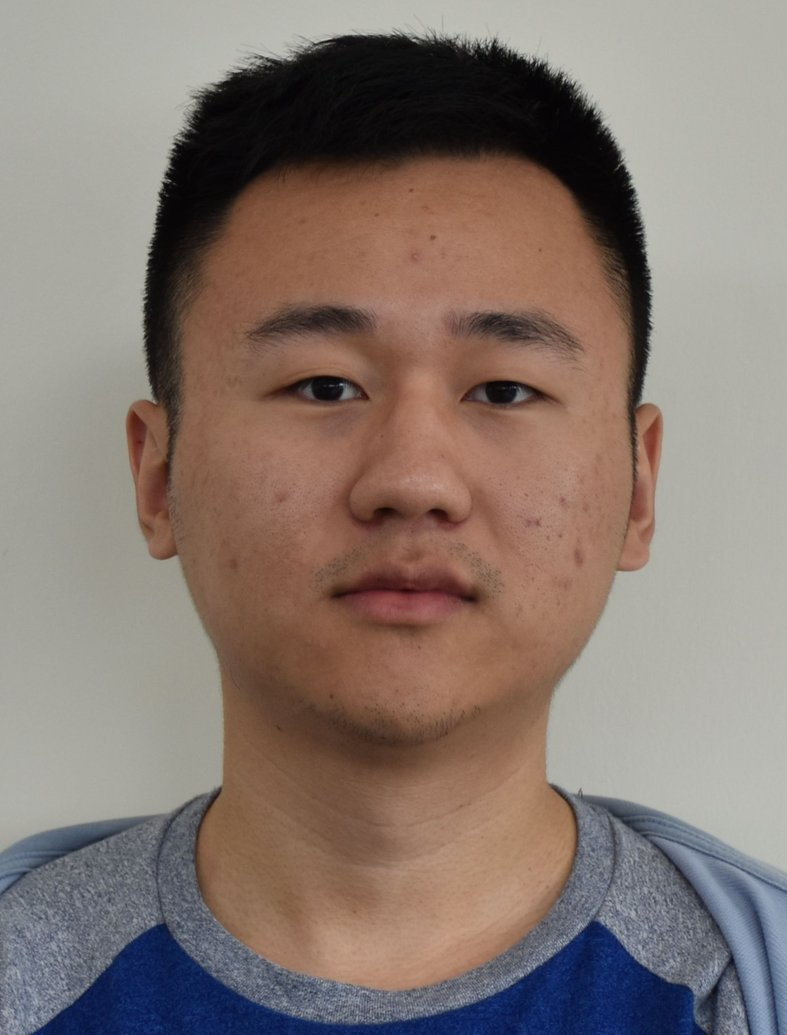}}]
{\textbf{Di Kang}} received the B.Eng. degree in Electronic Engineering and Information Science (EEIS) from the University of Science and Technology of China (USTC) in 2014, and the Ph.D. degree in computer science from City University of Hong Kong in 2019.
In 2019, he jointed Tencent AI Lab, and is now serving as a senior researcher.
His research interests include computer vision, and deep learning.
\end{IEEEbiography}

\vspace{-1.7cm}

\begin{IEEEbiography}[{\includegraphics[width=0.95\textwidth]{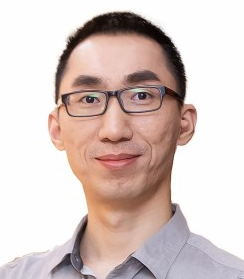}}]
{\textbf{Linchao Bao}} is currently a principal research scientist at Tencent AI Lab.
He received Ph.D. degree in Computer Science from City University of Hong Kong in 2015. Prior to that, he received M.S. degree in Pattern Recognition and Intelligent Systems from Huazhong University of Science and Technology in Wuhan, China. He was a research intern at Adobe Research from November 2013 to August 2014 and worked for DJI as an algorithm engineer from January 2015 to June 2016. His research interests include computer vision and graphics.
\end{IEEEbiography}

\vspace{-1.3cm}

\begin{IEEEbiography}[{\includegraphics[width=0.95\textwidth]{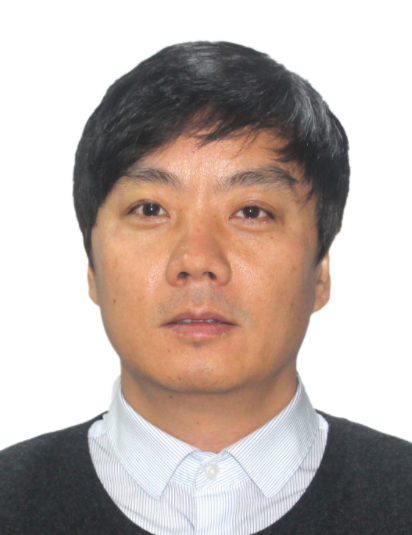}}]
{\textbf{Bisheng Yang}} received the B.S. degree in engineering survey, the M.S. and Ph.D. degrees in photogrammetry and remote sensing from Wuhan University, Wuhan, China, in 1996, 1999, and 2002, respectively. From 2002 to 2006, he was a Post-doctoral Research Fellow with the University of Zurich, Zurich, Switzerland. Since 2007, he has been a Professor with the State Key Laboratory of Information Engineering in Surveying, Mapping, and Remote Sensing, Wuhan University, where he is currently the Director of the 3S and Network Communication Laboratory. He has hosted a project of the National High Technology Research and Development Program, a key project of the Ministry of Education, and four National Scientific Research Foundation Projects of China. His current research interests include 3-D geographic information systems, urban modeling, and digital city. He was a Guest Editor of the ISPRS Journal of Photogrammetry and Remote Sensing and Computers \& Geosciences.
\vspace{0.14cm}
\end{IEEEbiography}

\vspace{-1.47cm}

\begin{IEEEbiography}[{\includegraphics[width=0.95\textwidth]{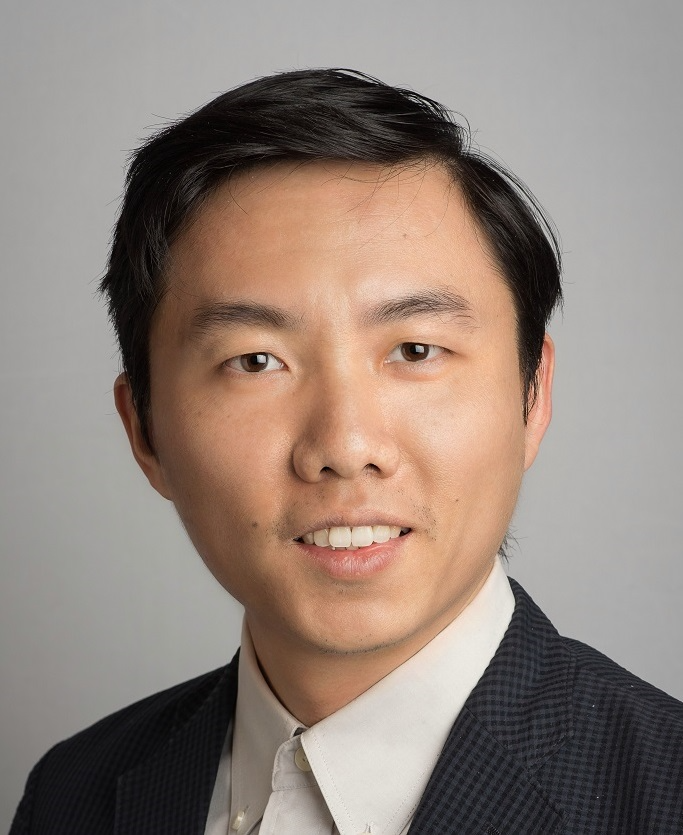}}]
{\textbf{Junsong Yuan}} is Professor and Director of Visual Computing Lab at Department of Computer Science and Engineering (CSE), State University of New York at Buffalo, USA. Before joining SUNY Buffalo, he was Associate Professor (2015-2018) and Nanyang Assistant Professor (2009-2015) at Nanyang Technological University (NTU), Singapore. He obtained his Ph.D. from Northwestern University in 2009, M.Eng. from National University of Singapore in 2005, and B.Eng. from Huazhong University of Science Technology (HUST) in 2002. His research interests include computer vision, pattern recognition, video analytics, human action and gesture analysis, large-scale visual search and mining. He received Best Paper Award from IEEE Trans. on Multimedia, Nanyang Assistant Professorship from NTU, and Outstanding EECS Ph.D. Thesis award from Northwestern University. He served as Associate Editor of IEEE Trans. on Image Processing (T-IP), IEEE Trans. on Circuits and Systems for Video Technology (T-CSVT), Machine Vision and Applications (MVA), and Senior Area Editor of Journal of Visual Communications and Image Representation (JVCI). He was Program Co-Chair of IEEE Conf. on Multimedia Expo (ICME'18), and served as Area Chair for CVPR, ICCV, and ACM MM. He is a Fellow of IEEE and IAPR.
\end{IEEEbiography}

\end{document}

%\linenumbers
\title{
\begin{Large}
\textbf{SUPPLEMENTARY MATERIAL}
\end{Large} \\
Consistent 3D Hand Reconstruction in Video via Self-Supervised Learning}
\markboth{IEEE Transactions on Pattern Analysis and Machine Intelligence}%
{Shell \MakeLowercase{\textit{et al.}}: Bare Demo of IEEEtran.cls for Computer Society Journals}

\author{Zhigang Tu$^*$,~\IEEEmembership{Member,~IEEE,}
        Zhisheng Huang$^*$,
        Yujin Chen$^\dag$, %~\IEEEmembership{Member,~IEEE,} Yujin is not a (student) member
        Di Kang,
        Linchao~Bao,~\IEEEmembership{Member,~IEEE,}
        Bisheng~Yang,~\IEEEmembership{Senior Member,~IEEE,}
        and Junsong Yuan,~\IEEEmembership{Fellow,~IEEE}}

\maketitle
%\appendix
\setcounter{table}{9}
\setcounter{figure}{11}

%\IEEEdisplaynontitleabstractindextext

% \fi
%\IEEEpeerreviewmaketitle

This is the appendix of the main text. 
In Section~\ref{sec:weights}, we demonstrate how to find the suitable loss weights and test how sensitive the results are to these loss weights. 
In Section~\ref{sec:sampling}, we show the stability of ${\rm {S}^{2}HAND(V)}$ sampling strategy during training. 
In Section~\ref{sec:TS}, we provide visualization of per-face texture standard deviation (S.D.).

\section{Additional Notes on Loss Weights}\label{sec:weights}

\subsection{Weights Selection}
\label{sec:selection}
In this section, we demonstrate how to find the suitable loss weights. 
% Overall,
In general, we find suitable initial weights by overfitting on a small dataset 
and further refine them through a divide-and-conquer grid search.

To obtain proper weights for multiple losses during training, we overfit our model on a small training set and select the weights based on the overfitting performances.
Take the weight initialization for $E_{geo}$ and $E_{regu}$ as an example. We first determine weighting factors in $E_{geo}$, i.e. weights for $E_{loc}$ and $E_{ori}$ ($\phi_{loc}$ and $\phi_{ori}$). 
We set the weight for $E_{loc}$ ($\phi_{loc}$) equal to 1, and find the best weight for $E_{ori}$ ($\phi_{ori}$). The initial weight ($\phi_{ori}$) is set to 1000 to unify the magnitudes of the weighted losses ($\phi_{loc}E_{loc}$ and $\phi_{ori}E_{ori}$). 
Then, we keep increasing or decreasing the weights by a factor of 10 until the sum of the weighted losses no longer decreases and the overfitting performance no longer improves. 
The sum of weighted losses only includes losses whose weights have been determined (i.e. $\phi_{loc}E_{loc}$ in this case). 
Finally, the weight combination that performs best is selected. 
We select 100 for the weight of $E_{ori}$ ($\phi_{ori}$) according to Table~\ref{table:ori}. 

\par
After a weight is selected, it is fixed for the remaining weight selections. 
Next, we determine weighting factors in $E_{regu}$ (i.e. weights for $E_{\beta}$, $E_{C}$, $E_s$, and $E_{J}$). 
Here, we only take the weight selection for $E_{J}$ as an example, and the rest are selected in the same manner. 
With the determined weights ($\phi_{loc}$ and $\phi_{ori}$) fixed, the initial weight of $E_{J}$ ($\phi_{J}$) is set to 1000 to unify the magnitudes of the weighted loss ($\phi_{loc}E_{loc}$ and $\phi_{J}E_{J}$). 
Then we explore the weight with a factor of 10 to find the best. According to Table~\ref{table:tsa}, 100 is selected. 
Repeating the above method, all the initialization weights can be determined. 

\par
Those overfitting experiments are conducted on 30 samples from the HO-3D dataset \cite{hampali2020honnotate} for 10 epochs. We use Adam optimizer with $\beta_1 = 0.9$ $\beta_2 = 0.999$. The batch size is set to 10 and the learning rate is set to 0.001. Each round of the experiment takes around 4 minutes.

\begin{table*}[ht]\color{black}
\begin{minipage}{1\linewidth}
\centering
\small
{\def\arraystretch{1} \tabcolsep=0.55em 
\caption{\textcolor{black}{Sensitivity of the selected weights. $\Omega_{X}$ denotes the selected weight of $E_{X}$ used in the main document.}}
\vspace{-0.3cm}
\label{table:sensitivity}
\scalebox{0.9}{
\begin{tabular}{cccccccccccc|cc|cccc}
\hline
       \multicolumn{12}{c|}{Ratio to the selected weights} & \multirow{2}{*}{$\rm{AUC}_{J}$$\uparrow$} & \multirow{2}{*}{MPJPE$\downarrow$} & \multirow{2}{*}{$\rm{AUC}_{V}$$\uparrow$} & \multirow{2}{*}{MPVPE$\downarrow$} & \multirow{2}{*}{${\rm F}_{5}$$\uparrow$}  & \multirow{2}{*}{${\rm F}_{15}$$\uparrow$} \\ \cline{1-12}
$\Omega_{loc}$ & $\Omega_{\beta}$ & $\Omega_{C}$ & $\Omega_{s}$ & $\Omega_{J}$ & $\Omega_{ori}$ & $\Omega_{2d}$ & $\Omega_{cons}$ & $\Omega_{pixel}$ & $\Omega_{SSIM}$ & $\Omega_{quat}$ & $\Omega_{T\&S}$ &        &       &        &       &       &       \\ \hline
1      & 1       & 1    & 1    & 1    & 1      & 1     & 1       & 1        & 1       & 0       & 0       & 0.755  & 1.236       &  0.760      &   1.207    &  0.439     &  0.918     \\ 
1      & 1       & 1    & 1    & 1    & 1      & 1     & 1       & 1        & 1       & 1       & 1       & 0.778  & 1.150 & 0.778  & 1.115 & 0.459 & 0.934 \\ \hline
$1/2$    & 1       & 1    & 1    & 1    & 1      & 1     & 1       & 1        & 1       & 1       & 1       & 0.769  & 1.158 & 0.777  & 1.118 & 0.459 & 0.930 \\ 
2      & 1       & 1    & 1    & 1    & 1      & 1     & 1       & 1        & 1       & 1       & 1       & 0.769  & 1.159 & 0.774  & 1.135 & 0.452 & 0.928 \\ \hline
1      & $1/2$     & 1    & 1    & 1    & 1      & 1     & 1       & 1        & 1       & 1       & 1       & 0.770  & 1.154 & 0.777  & 1.118 & 0.459 & 0.929 \\ 
1      & 2       & 1    & 1    & 1    & 1      & 1     & 1       & 1        & 1       & 1       & 1       & 0.770  & 1.151 & 0.777  & 1.120 & 0.459 & 0.932 \\ \hline
1      & 1       & $1/2$ & 1    & 1    & 1      & 1     & 1       & 1        & 1       & 1       & 1       & 0.769  & 1.166 & 0.773  & 1.137 & 0.455 & 0.931 \\ 
1      & 1       & 2    & 1    & 1    & 1      & 1     & 1       & 1        & 1       & 1       & 1       & 0.768  & 1.161 & 0.774  & 1.133 & 0.456 & 0.932 \\ \hline
1      & 1       & 1    & $1/2$  & 1    & 1      & 1     & 1       & 1        & 1       & 1       & 1       & 0.771  & 1.151 & 0.777  & 1.119 & 0.461 & 0.930 \\ 
1      & 1       & 1    & 2    & 1    & 1      & 1     & 1       & 1        & 1       & 1       & 1       & 0.769  & 1.159 & 0.776  & 1.124 & 0.456 & 0.931 \\ \hline
1      & 1       & 1    & 1    & $1/2$  & 1      & 1     & 1       & 1        & 1       & 1       & 1       & 0.772  & 1.146 & 0.776  & 1.124 & 0.453 & 0.930 \\ 
1      & 1       & 1    & 1    & 2    & 1      & 1     & 1       & 1        & 1       & 1       & 1       & 0.758  & 1.213 & 0.767  & 1.166 & 0.444 & 0.925 \\ \hline
1      & 1       & 1    & 1    & 1    & $1/2$   & 1     & 1       & 1        & 1       & 1       & 1       & 0.770  & 1.151 & 0.778  & 1.113 & 0.459 & 0.934 \\ 
1      & 1       & 1    & 1    & 1    & 2      & 1     & 1       & 1        & 1       & 1       & 1       & 0.768  & 1.164 & 0.775  & 1.129 & 0.455 & 0.928 \\ \hline
1      & 1       & 1    & 1    & 1    & 1      & $1/2$   & 1       & 1        & 1       & 1       & 1       & 0.769  & 1.159 & 0.776  & 1.125 & 0.457 & 0.931 \\ 
1      & 1       & 1    & 1    & 1    & 1      & 2     & 1       & 1        & 1       & 1       & 1       & 0.771  & 1.147 & 0.777  & 1.117 & 0.461 & 0.933 \\ \hline
1      & 1       & 1    & 1    & 1    & 1      & 1     & $1/2$     & 1        & 1       & 1       & 1       & 0.772  & 1.146 & 0.778  & 1.114 & 0.461 & 0.933 \\ 
1      & 1       & 1    & 1    & 1    & 1      & 1     & 2       & 1        & 1       & 1       & 1       & 0.765  & 1.178 & 0.771  & 1.147 & 0.453 & 0.926 \\ \hline
1      & 1       & 1    & 1    & 1    & 1      & 1     & 1       & $1/2$      & 1       & 1       & 1       & 0.769  & 1.159 & 0.775  & 1.126 & 0.459 & 0.932 \\ 
1      & 1       & 1    & 1    & 1    & 1      & 1     & 1       & 2        & 1       & 1       & 1       & 0.669  & 1.686 & 0.675  & 1.651 & 0.356 & 0.862 \\ \hline
1      & 1       & 1    & 1    & 1    & 1      & 1     & 1       & 1        & $1/2$     & 1       & 1       & 0.768  & 1.167 & 0.774  & 1.138 & 0.461 & 0.932 \\ 
1      & 1       & 1    & 1    & 1    & 1      & 1     & 1       & 1        & 2       & 1       & 1       & 0.772  & 1.146 & 0.777  & 1.119 & 0.461 & 0.932 \\ \hline
1      & 1       & 1    & 1    & 1    & 1      & 1     & 1       & 1        & 1       & $1/2$     & 1       & 0.757  & 1.221 & 0.764  & 1.187 & 0.440 & 0.918 \\ 
1      & 1       & 1    & 1    & 1    & 1      & 1     & 1       & 1        & 1       & 2       & 1       & 0.772  & 1.144 & 0.778  & 1.113 & 0.462 & 0.932 \\ \hline
1      & 1       & 1    & 1    & 1    & 1      & 1     & 1       & 1        & 1       & 1       & $1/2$     & 0.771  & 1.152 & 0.776  & 1.122 & 0.458 & 0.931 \\ 
1      & 1       & 1    & 1    & 1    & 1      & 1     & 1       & 1        & 1       & 1       & 2       & 0.766  & 1.175 & 0.772  & 1.144 & 0.450 & 0.927 \\ \hline
\end{tabular}}

}
\end{minipage}
\end{table*}

\begin{table*}[ht]\color{black}
\begin{minipage}{1\linewidth}
\centering
\small
{\def\arraystretch{1} \tabcolsep=2.1em 
\caption{\textcolor{black}{Weight Selection for $E_{ori}$. $\phi_{X}$ denotes the weight of $E_{X}$. The sum of weighted losses only includes losses whose weights have been determined, thus $\phi_{loc}E_{loc}$ is excluded. }}
\vspace{-0.3cm}
\label{table:ori}
\begin{tabular}{cc|ccc|cc}
\hline
\multicolumn{2}{c|}{Weights} & \multirow{2}{*}{$\phi_{loc}E_{loc}$} & \multirow{2}{*}{$\phi_{ori}E_{ori}$} & \multirow{2}{*}{\makecell{Weighted Loss Sum$\downarrow$\\ ($\phi_{loc}E_{loc}$)}} & \multirow{2}{*}{MPJPE$\downarrow$} & \multirow{2}{*}{$\rm {AUC}_{J}$$\uparrow$} \\ \cline{1-2}
$\phi_{loc}$       & $\phi_{ori}$        &                         &                         &                                &                        &                         \\ \hline
1            & 10           & 12.313                & 0.260                & 12.313                       & 1.242                  & 0.754                   \\ 
1            & 100          & 10.435                & 1.436                &  \underline{\textbf{10.435}}                       & \underline{\textbf{1.015}}                  &  \underline{\textbf{0.800}}                   \\ 
1            & 1000         & 13.027                & 13.881                & 13.027                    & 1.241                  & 0.752                   \\ 
1            & 10000        & 26.325                & 168.093                & 26.325                       & 1.647                  & 0.673                   \\ \hline
\end{tabular}}
\end{minipage}
\end{table*}

\begin{table*}[ht]\color{black}
\begin{minipage}{1\linewidth}
\centering
\small
{\def\arraystretch{1} \tabcolsep=1.4em 
\caption{\textcolor{black}{Weight Selection for $E_{J}$. $\phi_{X}$ denotes the weight of $E_{X}$. The sum of weighted losses only includes losses whose weights have been determined, thus $\phi_{J}E_{J}$ is excluded. }}
\vspace{-0.3cm}
\label{table:tsa}
\begin{tabular}{ccc|cccc|cc}
\hline
\multicolumn{3}{c|}{Weights} & \multirow{2}{*}{$\phi_{loc}E_{loc}$} & \multirow{2}{*}{$\phi_{ori}E_{ori}$} & \multirow{2}{*}{$\phi_{J}E_{J}$} & \multirow{2}{*}{\makecell{Weighted Loss Sum$\downarrow$ \\ ($\phi_{loc}E_{loc} + \phi_{ori}E_{ori}$)}} & \multirow{2}{*}{MPJPE$\downarrow$} & \multirow{2}{*}{$\rm {AUC}_{J}$$\uparrow$} \\ \cline{1-3}
$\phi_{loc}$   & $\phi_{ori}$   & $\phi_{J}$    &                         &                         &                       &                                &                        &                         \\ \hline
1        & 100     & 10     & 11.178                & 1.916                & 0.229              & 13.094                       & 1.017                  & 0.799                   \\
1        & 100     & 100    & 9.830                & 2.119                & 1.596              & \underline{\textbf{11.949}}                       & \underline{\textbf{0.927}}                  & \underline{\textbf{0.812}}                   \\ 
1        & 100     & 1000   & 12.035                & 1.556                & 4.079              & 13.591                       & 1.107                  & 0.779                   \\
1        & 100     & 10000  & 13.632                & 1.374                & 14.670               & 15.006                       & 1.243                  & 0.750                   \\ \hline
\end{tabular}}
\end{minipage}
\end{table*}

To further refine the selected weights, we perform a divide-and-conquer grid search. Weights within the same loss group constitute an individual searching space for ordinary grid search. Candidates of each weight can be the weight initialization multiplied by a factor less than 10 (e.g. a factor of 5, 3, or 2, etc.). 
${\rm {S}^{2}HAND}$ has 5 loss groups: $E_{loc}$ (1 sub-loss), $E_{regu}$ (4 sub-losses), $E_{ori}$ (1 sub-loss), $E_{2d}$ \& $E_{cons}$ (2 sub-losses) and $E_{photo}$ (2 sub-losses). When the number of candidates for one weight equals 3, the total number of experiments will be 105 (i.e. $3+3^{4}+3+3^{2}+3^{2}$). 
${\rm {S}^{2}HAND(V)}$ has 2 additional loss groups: $E_{quat}$ (1 sub-loss) and $E_{T\&S}$ (1 sub-loss), so the total number of experiments will be 111 (i.e. $105+3^{1}+3^{1}$).
Each round of experiment takes less than 5 minutes on the overfitting dataset, the full refinement can be done within 10 hours. In practice, we found that further refinement of some initialization weights did not improve much. Therefore, we empirically pruned the searching space during the refinement and the best weights we found have been presented in the main document. However, for those who want to push the envelope, the whole refinement space search is feasible and recommended.

\begin{figure*}[t]
\vspace{-0.1cm}
\centering
\includegraphics[width=\linewidth]{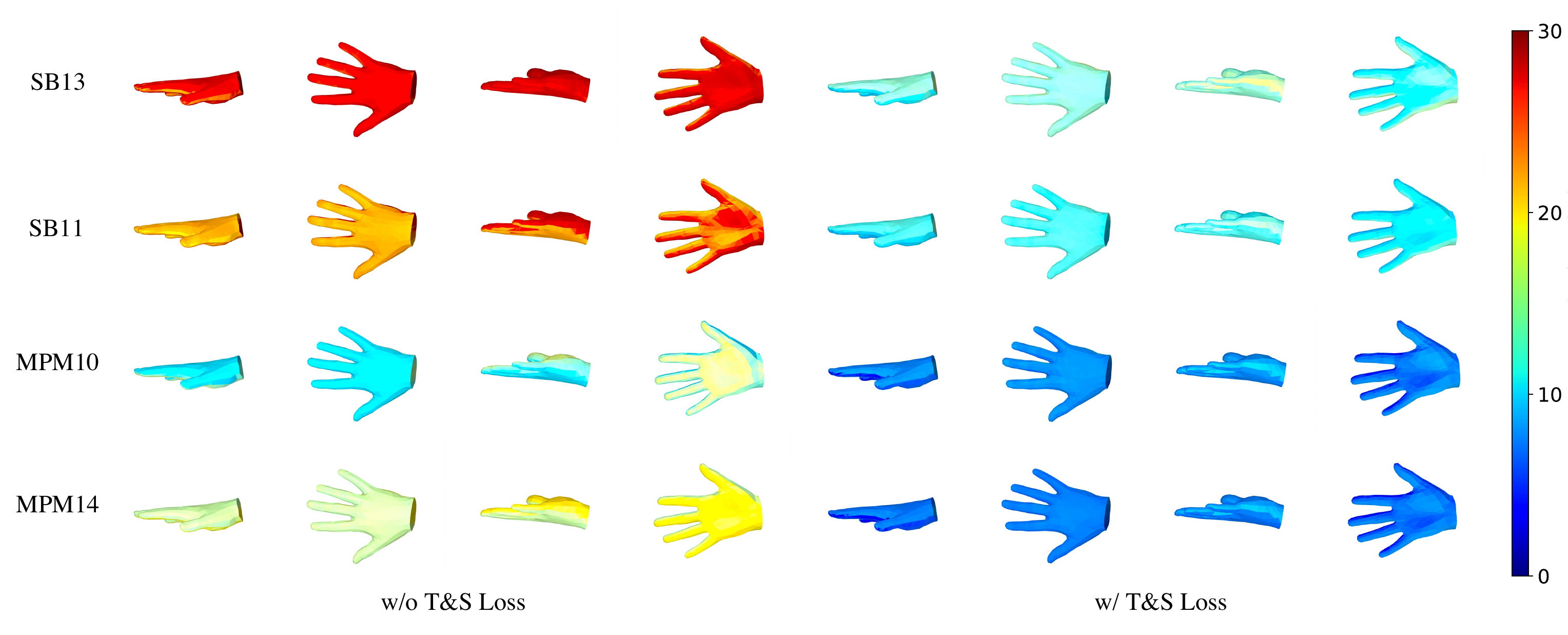}
\caption{\textcolor{black}{Visualization of the per-face texture S.D. in sequence predictions. 
SB13, SB11, MPM10 and MPM14 are sequences from the HO-3D \cite{hampali2020honnotate} testing set. 
The per-face texture S.D. are measured in RGB values (0-255) and presented in 4 different views (Column 1-4 and Column 5-8 are results w/o $\rm T\&S$ and w/ $\rm T\&S$ respectively). 
The $\rm T\&S$ consistency loss significantly decreases texture S.D..}}
\label{fig:TS}
%\label{fig:onecol}
%\vspace{-0.1cm}
\end{figure*}

\begin{table*}[ht]\color{black}
\begin{minipage}{\linewidth}
\centering
\small
{\def\arraystretch{1} \tabcolsep=0.7em 
\caption{\textcolor{black}{Results of repetitive experiments. ${\rm {S}^{2}HAND(V)}$ with the sequence sampling strategy gets stable results.}}
\vspace{-0.3cm}
\label{table:Stability of SS}
\begin{tabular}{c|cc|cc|cc|cc}
\hline
\multirow{3}{*}{Repetitions} & \multicolumn{4}{c|}{MPJPE} & \multicolumn{4}{c}{$\rm {AUC}_{J}$} \\ 
\cline{2-9} %\cline{8-12}
   & \multicolumn{2}{c|}{Test set} & \multicolumn{2}{c|}{Validation set} & \multicolumn{2}{c|}{Test Set} & \multicolumn{2}{c}{Validation Set}  \\ 
   \cline{2-9} %\cline{2-3} \cline{5-6}  \cline{8-9} \cline{11-12} 
   & ${\rm {S}^{2}HAND}$      & ${\rm {S}^{2}HAND(V)}$      & ${\rm {S}^{2}HAND}$         & ${\rm {S}^{2}HAND(V)}$    & ${\rm {S}^{2}HAND}$      & ${\rm {S}^{2}HAND(V)}$      & ${\rm {S}^{2}HAND}$         & ${\rm {S}^{2}HAND(V)}$  \\ 
   \hline
1  & 1.143664    & 1.100517       & 1.236116       & 1.152090     &     0.771680        &      0.780270          &     0.754652           &      0.770293  \\ 
2  & 1.145062    & 1.102492       & 1.237049       & 1.150140     &     0.771222       &       0.780027         &      0.754343          &       0.770780  \\ 
3  & 1.141907    & 1.097885       & 1.237436       & 1.153761     &     0.772158        &      0.780930          &             0.754180   &      0.769846  \\ 
\hline
\end{tabular}}
\end{minipage}
\end{table*}

\subsection{Weights Sensitivity}
\label{sec:ws}
We evaluate the sensitivity of our method to all the weighting factors. For one weighting factor, we give the evaluation results of changing its default value (presented in the main document) by multiplying it by 2 or 0.5 while keeping other weights unchanged.
Experiments are conducted on a train-validation split of the HO-3D dataset. 10  of 55 sequences in the training set are used as the validation set (GSF10, GSF13, SB10, GPMF11, SS1, ABF14, SiBF10, SiBF13, SiBF12, MC5). The rest 45 sequences are used for training.
As shown in Table~\ref{table:sensitivity}, for most of the weights, changing the value by multiplying it by 2 or 0.5 does not have a significant effect. However, doubling the weights of $E_{pixel}$ significantly degrades the performance. Examining the weighted loss ($\Omega_{pixel}E_{pixel}=0.009187$, $\Omega_{loc}E_{loc}=0.005935$) in the early training stage, we found that doubling the weight prioritizes texture learning over geometry learning. 
Thus the conclusion is that within the order of magnitude determined in the weight selection process, the weights are robust unless texture learning has seriously interfered with geometry learning.

\section{Stability of Sampling Strategy}\label{sec:sampling}
We use PyTorch\cite{paszke2019pytorch} framework to implement the ${\rm {S}^{2}HAND(V)}$ sampling strategy and test its stability through repetitive experiments. 
For the sampling, we first preprocess the raw dataset to obtain a sequence list and a list of sorted frames for each sequence. Then, a custom sampler \footnote{\href{https://pytorch.org/docs/stable/data.html?highlight=sampler\#torch.utils.data.Sampler}{https://pytorch.org/docs/stable/data.html}} 
% of
provided to our DataLoader is defined to do the training batch sampling. 
For each training batch, the custom sampler will draw $m$ sequences from the sequence list and sample $n$ frames for each of the $m$ sequence from the corresponding list of sorted frames. 
As mentioned in the main document, $m$ is set to $64 // n$, where 64 follows the batch size in training ${\rm {S}^{2}HAND}$ and $//$ represents floor division. The default value for $n$ is 3 (resulting in $m=21$), since it gives the best performance (see experimental results in Section~4.5.2 in the main document).
To verify the stability of our sampling strategy, we repeat ${\rm {S}^{2}HAND}$ and ${\rm {S}^{2}HAND(V)}$ experiments on both HO-3D train-test split and our train-validation split (Split the training set as described in Section~\ref{sec:ws}). 
As presented in Table~\ref{table:Stability of SS}, three rounds of experiments give similar results. 
${\rm {S}^{2}HAND(V)}$ and ${\rm {S}^{2}HAND}$ have similar and small performance fluctuations in the repetitive experiments. 
We attribute this to that our sampling strategy is consistent with the ordinary mini-batch sampling strategy in terms of randomness. 
To be specific, for both sampling strategies, given a certain number of iterations $I$ with dataset size $D$ and mini-batch size $B$, the mathematical expectation for the counts of drawing a specific sample is close to $\frac{IB}{D}$.
In conclusion, our ${\rm {S}^{2}HAND(V)}$ sampling strategy gets stable results.

\section{Visualization of Per-face Texture S.D.}
Fig.~\ref{fig:TS} visualizes the per-face texture S.D. in sequence predictions for four sequences.
It can be seen that the proposed $\rm T\&S$ consistency loss leads to a significant decrease of per-face texture S.D., resulting in more consistent sequence predictions. 
\label{sec:TS}

\bibliographystyle{IEEEtran}

\bibliography{hzsbib.bib}
